\documentclass[letterpaper]{article} 
\usepackage{aaai2026}  
\usepackage{times}  
\usepackage{helvet}  
\usepackage{courier}  
\usepackage[hyphens]{url}  
\usepackage{graphicx} 
\usepackage{natbib}  
\usepackage{caption} 
\usepackage[font=small]{subcaption}
\usepackage{algorithm}
\usepackage[noend]{algpseudocode} 
\usepackage{booktabs}   

\usepackage{newfloat}
\usepackage{listings}
\DeclareCaptionStyle{ruled}{labelfont=normalfont,labelsep=colon,strut=off} 
\floatstyle{ruled}
\newfloat{listing}{tb}{lst}{}
\floatname{listing}{Listing}

\usepackage{xspace}

\newcommand{\ourmodel}{VLM‑OpenXpert\xspace}

\usepackage{amsmath}  
\usepackage{amssymb}  
\usepackage{booktabs}
\usepackage{multirow}
\usepackage[table]{xcolor}     
\definecolor{tablegray}{gray}{0.90}

\title{Beyond Retraining: Training-Free Unknown Class Filtering for Source-Free Open Set Domain Adaptation of Vision–Language Models}
\author{
    Yongguang Li\textsuperscript{\rm 1},
    Jindong Li\textsuperscript{\rm 4},
    Qi Wang\textsuperscript{\rm 1,\rm2},
    QianLi Xing\textsuperscript{\rm 3},
    Runliang Niu\textsuperscript{\rm 1},
    Shengsheng Wang\textsuperscript{\rm 3},
    Menglin Yang\textsuperscript{\rm 4}
}

\affiliations{
    \textsuperscript{\rm 1}School of Artificial Intelligence, Jilin University,\\
    \textsuperscript{\rm 2}Engineering Research Center of Knowledge-Driven Human-Machine Intelligence, Ministry of Education, China\\
    \textsuperscript{\rm 3}College of Computer Science and Technology, Jilin University, Changchun, Jilin,
130012, China.\\
    \textsuperscript{\rm 4}The Hong Kong University of Science and Technology (Guangzhou)\\
    \textsuperscript{\rm 1,2,3}
    \{ygli25, niurl19\}@mails.jlu.edu.cn, \{qiwang, qianlixing, wss\}@jlu.edu.cn\\
    \textsuperscript{\rm 4}jli839@connect.hkust-gz.edu.cn, menglin.yang@outlook.com

}

\begin{document}
\makeatletter
\def\copyright@on{} 
\makeatother
\maketitle

\begin{abstract}
Vision-language models (VLMs) have gained widespread attention for their strong zero-shot capabilities across numerous downstream tasks. 
However, these models assume that each test image’s class label is drawn from a predefined label set and lack a reliable mechanism to reject samples from emerging unknown classes when only unlabeled data are available. 
To address this gap, open-set domain adaptation methods retrain models to push potential unknowns away from known clusters. Yet, some unknown samples remain stably anchored to specific known classes in the VLM feature space due to semantic relevance, which is termed as \textit{Semantic Affinity Anchoring (SAA)}. Forcibly repelling these samples unavoidably distorts the native geometry of VLMs and degrades performance. 
Meanwhile, existing score‑based unknown detectors use simplistic thresholds and suffer from threshold sensitivity, resulting in sub‑optimal performance.
To address aforementioned issues, we propose \ourmodel, which comprises two training‑free, plug‑and‑play inference modules.
SUFF performs SVD on high-confidence unknowns to extract a low-rank ``unknown subspace". Each sample’s projection onto this subspace is weighted and softly removed from its feature, suppressing unknown components while preserving semantics. 
BGAT corrects score skewness via a Box–Cox transform, then fits a bimodal Gaussian mixture to adaptively estimate the optimal threshold balancing known-class recognition and unknown-class rejection.
Experiments on 9 benchmarks and three backbones (CLIP, SigLIP, ALIGN) under source-free OSDA settings show that our training-free pipeline matches or outperforms retraining-heavy state-of-the-art methods, establishing a powerful lightweight inference calibration paradigm for open-set VLM deployment.
\end{abstract}

\section{Introduction}
\label{sec:Introduction}

Recent advances in large-scale vision-language models (VLMs) such as CLIP~\cite{2021_ICML_CLIP} have demonstrated powerful zero-shot capabilities across diverse downstream tasks without task-specific fine-tuning, making them foundational for real-world applications~\cite{2024_tpami_VLM_survey}. However, such zero-shot predictions inherently assume a closed-world scenario, where test samples must belong to a predefined label space. This assumption is frequently violated as unlabeled classes inevitably emerge in real-world scenarios~\cite{2025_CVIU_ODAwVL}. Misclassifying these unknown samples as known classes can lead to severe consequences. For instance, autonomous driving systems may misclassify unseen construction machinery as normal vehicles, posing serious safety hazards~\cite{2024_IJCV_OSR_REAL}. Moreover, real-world deployments typically lack labeled data and involve significant domain shifts, presenting critical challenges for enhancing unknown-class rejection in VLMs.

\begin{figure}
    \centering
    \begin{subfigure}[b]{0.56\linewidth}
        \includegraphics[height=3.68cm]{figs/fig1a.pdf}
        \label{fig:fig1_a}
    \end{subfigure}
    \hspace{0.005\linewidth}
    \begin{subfigure}[b]{0.41\linewidth}
        \includegraphics[height=3.68cm]{figs/fig1b.pdf}
        \label{fig:fig1_b}
    \end{subfigure}
    \caption{Analysis of CLIP on Office‑Home. Left: top‑8 unknown classes with the highest anchor rates and their anchored known classes. Right: HOS (\%) on four domains versus threshold; the dashed line marks the optimum.
}

    \label{fig:fig1}
\end{figure}

Against this backdrop, models must retain discriminative power on known classes while reliably rejecting unknown samples using only unlabeled target data. This challenge has given rise to Open-Set Unsupervised Domain Adaptation (OSDA) and its variant, Source-Free OSDA (SF-OSDA)~\cite{2020_TNNLS_ODA_SURVEY}. Conventional single-modal OSDA methods jointly train on labeled source and unlabeled target data to transfer knowledge and enhance unknown-class detection. The general idea of such methods is to enforce a repulsion loss that pushes potential unknown samples away from known-class clusters~\cite{2019_CVPR_STA}. In contrast, SF-OSDA dispenses with source data at adaptation time, relying solely on a source-pretrained model while still applying the same separation objective~\cite{CVPR_2024_UPUK}. In the VLM era, many works have directly transplanted this “retraining + separation” paradigm into vision–language settings by using VLM backbones, aiming to establish clear decision boundaries between known and unknown classes~\cite{2024_UEO_ICML,2023_ICMLW_UOTA}.

However, this paradigm implicitly relies on the global separability assumption, that unknown classes should be semantically distant from all known classes. Due to inherent semantic affinity, many unknown classes naturally anchor to specific known-class clusters. 
We term this phenomenon as \emph{Semantic Affinity Anchoring (SAA)}. For example, in Office-Home~\cite{2017_CVPR_OfficeHome}, nearly all samples of the unknown class ``keyboard” fall into the known ``computer” cluster (Figure~\ref{fig:fig1}(left)), and over 50\% of unknown classes exhibit SAA in CLIP. 
As SAA reflects the true semantic relationships captured by vision–language pretraining,  naively pushing anchored samples away risks distorting the model’s semantic structure and harming performance.
In addition to the geometric mismatch, VLM-based unknown sample detection also suffers from threshold sensitivity issue caused by existing score-based strategies. As shown in Figure \ref{fig:fig1}(right), optimal thresholds vary significantly across datasets. 
Prior approaches often adopt simplistic strategies such as fixed values~\cite{2025_CVIU_ODAwVL}, mean scores~\cite{2022_ECCV_ROS_OSDA}, k-means clustering~\cite{CVPR_2024_UPUK}, or heuristic rules, which fail to model score distributions effectively and lead to suboptimal performance.

To address such issues, we propose \ourmodel, a training-free and label-free source-free OSDA framework  of vision–language models, featuring two plug-and-play modules.
Specifically, the first module, SUFF (SVD-Based Unknown-Class Feature Filtering), which adopts a directional filtering strategy instead of global repulsion. It begins by selecting high-confidence unknown samples using an uncertainty-based criterion, explicitly including those hard-to-detect cases affected by Semantic Affinity Anchoring (SAA). Then, it performs SVD~\cite{1936_Psa_SVD} on the centralized features of these samples to construct a low-rank subspace, where unknowns exhibit higher projection energy than knowns. Finally, all samples are softly attenuated based on their projection in this subspace, effectively suppressing unknown-class features while preserving the semantic geometry of the model and mitigating the effects of SAA.

Next, we propose BGAT (Box-Cox GMM-Based Adaptive Thresholding) to replace brittle fixed or heuristic thresholds. BGAT first applies a Box-Cox transformation on confidence scores to stabilize scale and skewness, then fits a bimodal Gaussian mixture in the transformed space. The midpoint between the two modes yields a near-optimal, data-adaptive threshold. These two modules augment VLMs with reliable open-set detection capabilities at inference time with minimal computational overhead and without requiring any labeled samples or retraining. 
Experimental results demonstrate that our \ourmodel, despite being entirely training-free, outperforms representative training-intensive OSDA methods. Also, it can serve as a plug-and-play enhancement to boost the performance of existing trainable Source-free OSDA methods. \textbf{Our main contributions are summarized as follows}:
\begin{itemize}
    \item We identify and validate \textbf{\textit{Semantic Affinity Anchoring (SAA)}} phenomenon, where unknown samples being semantically attracted to known classes. We further show that excessively pushing unknowns away from knowns can distort VLM’s inherently generalized feature space, thereby degrading its performance.
    \item We propose SUFF, a training-free and plug-and-play module that suppresses unknown-class features via selective filtering, alleviating SAA while preserving VLM semantic geometry. Also, we introduce BGAT, a lightweight adaptive thresholding strategy that estimates near-optimal thresholds, enhancing the discriminative ability of VLM on identifying known classes while reliably rejecting unknown samples.
    \item Experiments on 9 benchmarks and 3 VLM backbones (e.g., CLIP, SigLIP, ALIGN) show that our training-free method matches or outperforms training-heavy representative OSDA methods, establishing a strong training-free baseline for VLM-based open-set domain adaptation.
\end{itemize}

\section{Related Work}
\label{sec:Related-Work}
\textbf{Vision-language Models.} 
Vision–Language Models (VLMs)~\cite{2024_tpami_VLM_survey} learn a shared embedding space for images and texts by contrastively aligning large-scale image–caption pairs, enabling simple image–text similarity to power a wide range of downstream tasks without task-specific heads. Representative models such as CLIP, SigLIP, and ALIGN normally adopt a twin-tower architecture with separate vision and text encoders. Such models are trained on massive web-scale datasets and support zero-shot inference via promptable text queries. While this design offers strong cross-task generalization, it lacks a built-in mechanism to reject unseen classes at test time~\cite{2023_CVPR_CLIPN}.
To address this limitation, we propose an inference-time calibration procedure that can be seamlessly integrated into any frozen VLM. Our method requires no labels or additional training and equips the model with robust open-set recognition capability, enabling reliable rejection of unknown samples during deployment.

\textbf{Unsupervised Open-Set Domain Adaptation.} OSDA~\cite{2017_iccv_osda,2025_Arxiv_CLIP_DA_DG_SURVEY} aims to mitigate the decline in model performance caused by distribution discrepancies between the source and target domains, while assuming that the target domain contains unknown categories absent in the source domain. Traditional OSDA methods~\cite{2019_CVPR_STA,2022_ECCV_ROS_OSDA} typically rely on training with both source and target domain data concurrently. However, practical applications may restrict direct access to source domain data during target domain training due to privacy concerns, thereby motivating the development of Source-Free Unsupervised Domain Adaptation (SF-OSDA)~\cite{2023_tpami_PGL,2024_CVPR_LEAD}. Most existing SF-OSDA~\cite{2023_tpami_PGL,2024_CVPR_LEAD} approaches substitute source domain samples with models pre-trained on the source domain for target domain training. In recent years, capitalizing on CLIP’s remarkable zero-shot transfer capabilities, some SF-OSDA methods have begun to integrate CLIP with a set of known class names to train directly on unlabeled target domain data~\cite{2023_ICMLW_UOTA,2024_UEO_ICML}. This paper specifically addresses this scenario by proposing a method that requires no training.

\begin{figure*}[!t]
    \centering
    \includegraphics[width=0.99\linewidth]{figs/framework.pdf}
    \caption{Overview of VLM-OpenXpert, which performs lightweight, training-free inference on unlabeled target data via entropy-based scoring, unknown-class feature filtering (SUFF), and adaptive thresholding (BGAT).}
    \label{fig:framework}
\end{figure*}


\section{Methodology}
\label{sec:Methodology}

Given an unlabeled target dataset $D_t = \{x_i^t\}_{i=1}^{N_t}$ and a set of known categories $L_t = \{l_c\}_{c=1}^{C}$, where $l_c$ represents the category name of the $c$-th class, some samples in $D_t$ belong to $L_t$, while others do not. Our goal is to perform inference on $D_t$ by assigning samples belonging to $L_t$ to their respective known classes and classifying the rest as an unknown class $(C{+}1)$. 
The model’s performance is evaluated on $D_t$.

As shown in Figure~\ref{fig:framework}, VLM-OpenXpert uses a frozen VLM backbone for zero-shot inference to produce known-class predictions and compute per-sample scores. It then applies BGAT to estimate a threshold for identifying representative samples to construct the SUFF module, suppresses unknown-class features via SUFF, recomputes scores, and uses BGAT again to determine the threshold for final detection of unknown-class samples.

\subsection{Box-Cox GMM-Based Adaptive Thresholding (BGAT) Module}
\label{sec_Method_subsec_BGAT}

Traditional open-set domain adaptation methods typically employ fixed thresholds (e.g., Max/2) ~\cite{2025_CVIU_ODAwVL} or global means~\cite{2022_ECCV_ROS_OSDA,2019_CVPR_STA} to distinguish unknown-class samples, which lack theoretical foundation and oversimplify real-world distributions. To investigate a more universal and adaptive threshold calibration strategy, motivated by previous work~\cite{2024_CVPR_USD}, we introduce a statistically grounded approach combining Box-Cox transformation~\cite{1964_book_for_box_cox} and Gaussian Mixture Models (GMM)~\cite{2006_BOOK_FOR_GMM}.  

Specifically, as shown in Figure~\ref{fig:framework}, given an unlabeled target dataset $D_t$ and a set of known class names $L_t = \{l_1, l_2, \dots, l_C\}$, we perform zero-shot classification on all samples in $D_t$ using a VLM (e.g., CLIP). The classification probabilities for known classes and the detection of unknown samples are computed as follows:

\begin{equation}
\label{euqa:hat_y}
\hat{y}_{i,c} = \frac{\exp\left( \cos\left(X_{all}[i], X_{text}[c]\right) / t \right)}{\sum_{c'=1}^C \exp\left( \cos\left(X_{all}[i], X_{text}[c']\right) / t \right)},
\end{equation}
\begin{equation}
\text{Pred}(x_i) =
\begin{cases}
C+1, & \text{if } \text{Score}(\hat{y}_i) > T^*, \\
\arg\max_c \hat{y}_{i,c}, & \text{otherwise}.
\label{euqa:pred_by_T}
\end{cases}
\end{equation}

In particular, we define:
\begin{equation}
    X_{all} = \{E(x_i) \mid x_i \in D_t\}, \quad X_{text} = \{G(p(l_c)) \mid l_c \in L_t\},
    \label{equa:x_all_and_x_text}
\end{equation}
where $E$ denotes the image encoder, $G$ denotes the text encoder, and $p(l_c)$ is the class description sentence with the default prompt “a photo of a”. The threshold is denoted by $T^*$, and we adopt entropy as the default sample score:
\begin{equation}
    \text{S}(x_i) = H(\hat{y}_i) = -\sum_{c=1}^C \hat{y}_{i,c} \log(\hat{y}_{i,c}).
    \label{euqa:score_entropy}
\end{equation}

We assume that the score distributions for known-class and unknown-class samples follow Gaussian distributions with different means and variances. 
Under the common assumption of symmetric costs and comparable priors, the optimal threshold $T^*$ admits the well-known closed form given by the intersection of the two densities (see Theorem 1 in Appendix \ref{appendix:Proof}). 
However, there are two issues in practical datasets that need to be addressed:

\textit{$(i)$ Distribution Skewness}: 
Taking the entropy of the samples as an example, the entropy of known classes is concentrated near 0 due to the high certainty of most known-class samples. This results in a significant left skew in distribution, violating the assumption of a Gaussian distribution;
\textit{$(ii)$ Error in GMM Estimation}: During the GMM estimation process, errors arise, particularly when the entropy of the samples is highly concentrated, the variance estimation for known classes becomes notably inaccurate, significantly reducing the accuracy of intersection points.

Addressing Issue $(i)$: We apply the Box–Cox~\cite{1964_book_for_box_cox} transformation; to ensure positivity, add a tiny $\epsilon>0$ to the scores (still denoted $S_i$). The transform is:

\begin{equation}
    S_i^{(\lambda)} =
\begin{cases}
\dfrac{S_i^\lambda - 1}{\lambda}, & \lambda \neq 0,\\[4pt]
\log S_i, & \lambda = 0.
\end{cases}
\label{euqa:box_cox_trans}
\end{equation}

where the optimized parameter \(\lambda^*\) is estimated using maximum likelihood~\cite{1921_for_maxlikelihood} as:
\begin{equation}
\lambda^* = \arg\max_{\lambda}\;\;
\sum_{i=1}^{N} \log \mathcal{N}\!\big(S_i^{(\lambda)};\,\mu_\lambda,\sigma_\lambda^2\big)
\;+\; (\lambda-1)\sum_{i=1}^{N}\log S_i .
\label{eq:boxcox_mle}
\end{equation}

Since this data transformation is monotonically increasing, it does not alter the relative ordering of the sample scores. However, it can alleviate the issue of sample scores being too close to known classes, thereby improving the accuracy of mean and variance estimation.

Addressing Issue $(ii)$: After Box-Cox transformation, we fit a two-component GMM to $\{S_i^{(\lambda^*)}\}$ as:  
\begin{equation}
\label{euq:EM_GMM}
p\!\left(S^{(\lambda^*)}\right)=
\pi_1 \mathcal{N}(\mu_1,\sigma_1^2)+\pi_2 \mathcal{N}(\mu_2,\sigma_2^2),
\end{equation}

Parameters $\{\pi_k, \mu_k, \sigma_k\}_{k=1}^2$ are estimated via EM~\cite{1977_for_EM} and sorted as $(\mu_{\text{low}}, \sigma_{\text{known}}), (\mu_{\text{high}}, \sigma_{\text{unk}})$. 
However, the GMM estimation may introduce errors, particularly the underestimation of the known-class variance $\sigma_{\text{know}}$. Since the intersection point is highly sensitive to the variance (as shown in Theorem 2 of Appendix~\ref{appendix:Proof}), it may not be the optimal choice in practice.

To address this issue, we adopt a compromise approach by selecting the midpoint of the means, \( T_{trans}^* = \frac{\mu_{know} + \mu_{unk}}{2} \), as the threshold. This choice is motivated by the fact that the mean estimated by GMM is less sensitive to errors than the variance, and using the midpoint can enhance the robustness of the threshold. Finally, the optimal threshold \(T^*\) is derived via inverse Box-Cox transformation as:  

\begin{equation}
T^* =
\begin{cases}
\big(\lambda^* T_{\text{trans}}^* + 1\big)^{1/\lambda^*}, & \lambda^* \neq 0,\\[4pt]
\exp\!\big(T_{\text{trans}}^*\big), & \lambda^* = 0.
\end{cases}
\label{euqa:inverse_box_cox}
\end{equation}

In addition, the estimated means $(\mu_{\text{know}},\mu_{\text{unk}})$ of the known and unknown classes also need to be transformed back to the space before the Box-Cox transformation.

\subsection{SVD-Based Unknown-Class Feature Filtering (SUFF) Module}
Although the BGAT module estimates the optimal threshold $T^*$, the SAA phenomenon still causes some unknown‑class samples to have low entropy and be misclassified as known‑class samples.
To resolve this, we propose an SVD-based Unknown Feature Filtering Module (SUFF), which filters out unknown-class features from $X_{\text{all}}$ to mitigate the SAA phenomenon.

Specifically, we use the known-class mean $\mu_{\text{know}}$, along with the unknown-class mean $\mu_{\text{unk}}$ and threshold $T^*$, both estimated by the BGAT module from VLM zero-shot scores, to filter high-confidence known- and unknown-class sample features.
We filter samples with entropy less than or equal to $\frac{\mu_{\text{know}} + T^*}{2}$, forming the high-confidence known-class sample feature set $X_{\text{know}} \in \mathbb{R}^{N_{\text{know}} \times D}$; and we filter samples with entropy greater than$ \frac{\mu_{\text{unk}} + T^*}{2}$, forming the high-confidence unknown-class sample feature set $X_{\text{unk}} \in \mathbb{R}^{N_{\text{unk}} \times D}$.

Next, we perform SVD on both known-class and unknown-class feature sets to construct their principal component subspaces as:
\begin{equation}
\label{euqa:svd_x_know}
    X_{\text{know}} = U_{\text{know}} \Sigma_{\text{know}} V_{\text{know}}^\top,
\end{equation}
\begin{equation}
\label{svd_x_unk}
    X_{\text{unk}} = U_{\text{unk}} \Sigma_{\text{unk}} V_{\text{unk}}^\top,
\end{equation}
where $\Sigma_{\text{know}} = \text{diag}(\sigma_1^{\text{know}}, \dots, \sigma_r^{\text{know}})$ and $\Sigma_{\text{unk}} = \text{diag}(\sigma_1^{\text{unk}}, \dots, \sigma_r^{\text{unk}})$ are the singular value matrices. $V_{\text{know}} \in \mathbb{R}^{r \times D}$ and $V_{\text{unk}} \in \mathbb{R}^{r \times D}$ are the right singular matrices. The smallest dimension \(k\) is selected such that the cumulative variance contribution satisfies the following:

\begin{equation}
    \min k \quad \text{s.t.} \quad \frac{\sum_{i=1}^k (\sigma_i)^2}{\sum_{j=1}^r (\sigma_j)^2} \geq \tau.
    \label{equa:svd_tau}
\end{equation}

\begin{table*}[htbp]
\renewcommand{\arraystretch}{1.2}
  \centering
    \resizebox{\textwidth}{!}{
    \begin{tabular}{c|c|c|c|c|c|c|c|c|c|c|c|c|c|c|c|c|c}
    \toprule[2pt]
    \multirow{3}[0]{*}{method} & \multirow{3}[0]{*}{E} & \multirow{3}[0]{*}{VLM} & \multirow{3}[0]{*}{TF} & \multicolumn{13}{c|}{OfficeHome}                                                                       & VisDA \\ \cline{5-18}
          &       &       &       & C     & P     & R     & A     & P     & R     & A     & C     & R     & A     & C     & P     & \multirow{2}[0]{*}{AVG} & Syn \\ \cline{5-16} \cline{18-18}
          &       &       &       & \multicolumn{3}{c|}{A} & \multicolumn{3}{c|}{C} & \multicolumn{3}{c|}{P} & \multicolumn{3}{c|}{R} &       & Real \\ \cline{1-18}
    \multicolumn{18}{>{\columncolor{tablegray}}c}{OSDA (trains jointly on labeled source data and unlabeled target data)} \\ \cline{1-18}
    USD   & RN    & $\times$     & $\times$     & 60.1  & 62.6  & 67.8  & 61.1  & 56.3  & 59.1  & 70.0  & 65.2  & 71.1  & 76.3  & 68.9  & 56.3  & 64.6  & 69.4  \\
    OSMDA-CLIP & Vit-B-32 & $\checkmark$      & $\times$     & 74.1  & 74.7  & 76.6  & 71.9  & 69.7  & 70.4  & 81.9  & 82.2  & 81.8  & 83.8  & 84.0  & 69.7  & 76.7  & 83.4  \\
    UniOT-CLIP & Vit-B-16 & $\checkmark$      & $\times$     & 79.2  & 71.8  & 77.1  & 77.8  & 72.3  & 76.0  & 83.3  & 83.1  & 87.3  & 86.3  & 84.1  & 84.7  & 80.3  & - \\
    GLC-CLIP & Vit-B-16 & $\checkmark$      & $\times$     & 55.5  & 52.1  & 69.0  & 72.1  & 65.9  & 71.3  & 79.7  & 81.3  & 83.9  & 83.3  & 77.9  & 65.9  & 71.5  & 83.4  \\ \hline
    \multicolumn{18}{>{\columncolor{tablegray}}c}{SF-OSDA (uses a model trained on labeled source data and performs unsupervised training on unlabeled target data)} \\ \hline
    LEAD  & RN    & $\times$     & $\times$     & 61.0  & 65.5  & 64.8  & 60.7  & 59.8  & 57.7  & 70.8  & 68.6  & 75.8  & 76.5  & 70.8  & 59.8  & 66.0  & 74.2  \\
    DTDE  & RN    & $\times$     & $\times$     & -     & -     & -     & -     & -     & -     & -     & -     & -     & -     & -     & -     & 70.3  & 80.4  \\
    (Co-learn++)-CLIP & RN    & $\checkmark$      & $\times$     & 60.4  & 56.0  & 64.0  & 54.9  & 51.0  & 58.8  & 77.6  & 72.7  & 83.6  & 78.4  & 75.9  & 51.0  & 65.4  & - \\
    COSDA & RN    & $\times$     & $\times$     & 65.0  & 67.8  & 70.3  & 70.5  & 63.7  & 64.1  & 74.9  & 72.2  & 79.1  & 79.9  & 76.1  & 63.7  & 70.6  & - \\
    UPUK  & RN    & $\times$     & $\times$     & 66.4  & 67.6  & 67.8  & 55.8  & 55.1  & 59.4  & 76.7  & 73.1  & 74.4  & 78.4  & 77.6  & 55.1  & 67.3  & 72.3  \\
    COCA-CLIP & Vit-B-16 & $\checkmark$      & $\times$     & 79.7  & 79.6  & 80.0  & 75.6  & 74.5  & 75.7  & 84.5  & 84.3  & 84.4  & 82.5  & 82.5  & 74.5  & 79.8  & 86.3  \\
    COSDA-CLIP & Vit-B-32 & $\checkmark$      & $\times$     & -     & -     & -     & -     & -     & -     & -     & -     & -     & -     & -     & -     & -     & 78.4  \\
    DIFO-CLIP & Vit-B-32 & $\checkmark$      & $\times$     & 68.2  & 67.2  & 71.9  & 64.5  & 62.1  & 65.3  & 86.2  & 79.3  & 84.4  & 87.9  & 86.1  & 62.1  & 73.8  & - \\ \hline
    \multicolumn{18}{>{\columncolor{tablegray}}c}{Target-only Adaptation (adapts directly on the unlabeled target domain; notably, only our method performs inference without any training)} \\ \hline

    UOTA-CLIP & Vit-B-32 & $\checkmark$      & $\times$     & \multicolumn{3}{c|}{75.9  } & \multicolumn{3}{c|}{75.1 } & \multicolumn{3}{c|}{84.2 } & \multicolumn{3}{c|}{86.2 } & 80.4  & 85.3  \\
    UEO-CLIP & Vit-B-16 & $\checkmark$      & $\times$     & \multicolumn{3}{c|}{72.1 } & \multicolumn{3}{c|}{71.1 } & \multicolumn{3}{c|}{79.5 } & \multicolumn{3}{c|}{79.5 } & 75.6  & 82.6  \\
        ours-CLIP & RN    & $\checkmark$     & $\checkmark$     & \multicolumn{3}{c|}{73.5} & \multicolumn{3}{c|}{61.9} & \multicolumn{3}{c|}{81.9} & \multicolumn{3}{c|}{81.6} & 74.7   & 85.1   \\
    ours-CLIP & Vit-B-32 & $\checkmark$     & $\checkmark$     & \multicolumn{3}{c|}{75.1} & \multicolumn{3}{c|}{72.8} & \multicolumn{3}{c|}{85.1} & \multicolumn{3}{c|}{87.2} & 80.0  & 87.6  \\
    ours-CLIP & Vit-B-16 & $\checkmark$     & $\checkmark$     & \multicolumn{3}{c|}{78.2} & \multicolumn{3}{c|}{74.0} & \multicolumn{3}{c|}{86.8} & \multicolumn{3}{c|}{85.7} & 81.2   & \textbf{89.0} \\
    ours-SigLIP & Vit-B-16 & $\checkmark$     & $\checkmark$     & \multicolumn{3}{c|}{\textbf{85.8}} & \multicolumn{3}{c|}{\underline{82.9}} & \multicolumn{3}{c|}{\underline{90.0} } & \multicolumn{3}{c|}{\underline{89.2}} & \underline{87.0}  & \underline{88.3}  \\
    ours-ALIGN & EfficientNet-B7 & $\checkmark$     & $\checkmark$     & \multicolumn{3}{c|}{ \underline{85.0} } & \multicolumn{3}{c|}{\textbf{83.2}} & \multicolumn{3}{c|}{\textbf{90.9}} & \multicolumn{3}{c|}{\textbf{91.3}} & \textbf{87.6} & 87.3  \\

    \bottomrule[2pt]
    \end{tabular}%
    }

    \caption{HOS (\%) on Office-Home and VisDA-2017. Bold denotes the best results and underline denotes the second best. ``-CLIP", ``-SigLIP", and ``-ALIGN'' represent using the respective VLM backbone. (E: Image Encoder; SF-S: Source-Free; TF: Training-Free).}
\label{table:full_res_officehome_visda2017}
\end{table*}%

We retain the top $k$ principal components to construct the projection matrices $W_{\text{know}} = V_{\text{know}}[:k, :]^\top$ and $W_{\text{unk}} = V_{\text{unk}}[:k, :]^\top$. Then, we project all sample features  $X_{\text{all}} \in \mathbb{R}^{N \times D}$  into the constructed known class space  $Z_{know}$ and unknown class space  $Z_{unk}$ as follows:
\begin{equation}
    X_{\text{all}}^{\text{know}} = Z_{know}( X_{\text{all}}) =  X_{\text{all}} W_{\text{know}} W_{\text{know}}^\top,
    \label{euqa:proj_x_know}
\end{equation}
\begin{equation}
    X_{\text{all}}^{\text{unk}}= Z_{unk}( X_{\text{all}}) = X_{\text{all}} W_{\text{unk}} W_{\text{unk}}^\top.
    \label{euqa:proj_x_unk}
\end{equation}

Next, we analyze the proportion $\alpha(X_{\text{all}})$ of unknown class features in the samples based on the similarity $S(x_{\text{all}})$ before and after normalizing the sample feature mapping:
\begin{equation}
    s_{\text{know}}(X_{all}) = \frac{X_{all} \cdot X_{\text{all}}^{\text{know}}}{\|X_{all}\| \|X_{\text{all}}^{\text{know}}\|},
\end{equation}
\begin{equation}
    s_{\text{unk}}(X_{all}) = \frac{X_{all} \cdot X_{\text{all}}^{\text{unk}}}{\|X_{all}\| \|X_{\text{all}}^{\text{unk}}\|},
\end{equation}
\begin{equation}
    \alpha(X_{all}) = \frac{\exp(s_{\text{unk}}(X_{all}) / t)}{\exp(s_{\text{know}}(X_{all})/ t) + \exp(s_{\text{unk}}(X_{all}) / t)},
    \label{euqa:calculate_alpha}
\end{equation}
where \(t\) is temperature coefficient. \(t\) is set to 1.0 by default and it controls the smoothness of probability distribution.

To remove the unknown-class features from the sample features, we perform a weighted subtraction operation:
\begin{equation}
    X_{all}^* = X_{all} - \alpha(X_{all}) X_{\text{all}}^{\text{unk}},
    \label{euqa:sub_unkonw_feature}
\end{equation}

Finally, we recompute sample scores using the filtered features $X_{\text{all}}^*$ and $X_{\text{text}}$, and re-estimate the threshold $T^*$ via BGAT. Samples scoring above $T^*$ are marked as unknown, while others retain their original predictions from Eq.~\ref{euqa:hat_y}, preserving known-class recognition and mitigating filtering-induced errors. 

\section{Experiments}
\label{sec:Experiments}

\subsection{Experimental Settings}
\textbf{Datasets.} 
We evaluate our method under the SF‑OSDA experimental setting on three representative benchmark datasets: Office-Home~\cite{2017_CVPR_OfficeHome} (65 categories, 4 domains), VisDA-2017~\cite{2017_arXiv_VisDA} (12 categories, synthetic-to-real adaptation), and DomainNet~\cite{2019_ICCV_DomainNet} (345 categories, 6 domains). To further test the generalization ability of our model, we select 6 datasets from the Visual Task Adaptation Benchmark (VTAB-6). 

\textbf{Baselines.}
We select three categories of baseline methods for comparison:
\textbf{OSDA/SF-OSDA methods} jointly train on source data (or a source‐trained model) and unlabeled target data, including USD~\cite{2024_CVPR_USD}, LEAD~\cite{2024_CVPR_LEAD}, DIFO~\cite{CVPR_2024_DIFO}, OSMDA~\cite{TCSVT_2025_OSMDA}, UniOT~\cite{NIPS_2022_UniOT}, GLC~\cite{CVPR_2023_GLC}, COCA~\cite{ACCV_2024_COCA},  DTDE~\cite{AAAI_2025_DTDE}, Co-learn++~\cite{IJCV_2025_COLEARN}, COSDA~\cite{2025_ICML_COSDA}, UPUK~\cite{CVPR_2024_UPUK};
\textbf{Unsupervised target-only adaptation methods} such as UOTA~\cite{2023_ICMLW_UOTA} and UEO~\cite{2024_UEO_ICML}, which perform training and unknown class detection directly on the target domain; \textbf{CLIP-based zero-shot OOD methods} using Maximum Concept Matching (MCM)~\cite{2022_NIPS_MCM}. 

\begin{table}[htbp]
  \centering
    \renewcommand{\arraystretch}{1.2}
    \resizebox{\linewidth}{!}{
    \begin{tabular}{c|c|c|c|c}
           \toprule[1.5pt]
    Method & E     & VTAB-6 & DomainNet & AVG \\ \hline
    MCM-CLIP & ViT-L/14 & 59.9  & 67.8  & 63.9 \\
    MCM-SigLIP & ViT-B/16 & 66.3  & 66.5  & 66.4 \\
    MCM-ALIGN & EfficientNet-B7 & 59.3  & 66.6  & 62.9 \\
    UOTA-CLIP & ViT-L/14 & -     & 71.1  & - \\
    UEO-CLIP & ViT-L/14 & 57    & 69.6  & 63.3 \\
    ours-CLIP & ViT-L/14 & \underline{65.4}  & \textbf{72.9} & \underline{69.1} \\
    ours-SigLIP & ViT-B/16 & \textbf{70.2} & \underline{70.8}  & \textbf{70.5} \\
    ours-ALIGN & EfficientNet-B7 & 62.2   & 66.9   & 64.5  \\
           \bottomrule[1.5pt]
    \end{tabular}%
    }
      \caption{HOS (\%) on DomainNet and six visual task adaptation datasets.}
  \label{tab:vatb_domainnet_performance}%
\end{table}%


\textbf{Implementation Details.}
We evaluate on three VLM backbones—CLIP~\cite{2021_ICML_CLIP}, SigLIP~\cite{2023_ICCV_SigLIP}, and ALIGN~\cite{2021_ICML_ALIGN}. Image encoders follow the table headers; when unspecified, we default to ViT-B/16~\cite{2022_TPAMI_VIT} for CLIP and SigLIP, and EfficientNet-B7~\cite{2019_PMLR_EfficientNet} for ALIGN. Unless otherwise noted, we set $\tau=0.8$ and use a batch size of 32. See Appendix~\ref{appendix:Experiments-Datasets} for more details.

\subsection{Overall Performance}
As shown in Table~\ref{table:full_res_officehome_visda2017}  and Table~\ref{tab:vatb_domainnet_performance}, our training‑free approach surpasses previous OSDA, SF‑OSDA, and target‑only adaptation methods. Using the same backbone on Office‑Home, it outperforms GLC, DIFO, and UEO by 9.7\%, 7.4\%, and 5.6\%, respectively. Compared with the state‑of‑the‑art UOTA, it achieves gains of 3.7\% on VisDA‑2017 and 1.8\% on DomainNet, while matching UOTA on Office‑Home.
These results confirm the effectiveness of our method and show that an efficient, training‑free inference strategy can unlock the inherent open‑set zero‑shot capability of VLMs on unlabeled target data. Moreover, our approach consistently outperforms MCM on three different VLM backbones across all six Visual Task Adaptation benchmarks, demonstrating the robust improvements in unknown‑class detection. Notably, our method adds only 2.5\% inference time compared to direct zero-shot use of VLMs, while being tens of times faster than conventional training-intensive methods. See Appendices \ref{appendix:overall_performance} and \ref{appendix:time_analyse} for more details.

\subsection{Ablation Study}

\begin{figure}[]
    \centering
    \includegraphics[width=\linewidth]{figs/threshold_ablation.pdf}
    \caption{HOS(\%) performance variation under different thresholding methods.}
    \label{fig:threshold_ablation}
\end{figure}

\textbf{Analysis of the SAA Phenomenon.}
To quantitatively validate how much unknown samples concentrate around a single known class, we propose the Anchor Rate at 1 (e.g., AR@1) metric, which is defined as the fraction of unknown classes whose strongest semantic affinity is with one known class (e.g., range 0–1). As shown in Table~\ref{tab:AR@1}, all three VLMs exhibit high AR@1 values (0.46–0.58) on the widely used benchmarks, indicating that about half of unknown classes in these benchmarks anchor to specific known-class clusters, thereby confirming the widespread presence of the SAA phenomenon. See Appendix~\ref{appendix:Ablation-Study} for detailed analysis.

\textbf{Effectiveness of SUFF.}
 As shown in Table~\ref{tab:AR@1}, SUFF reduces AR@1 by 0.23, 0.28, and 0.33 on CLIP, SigLIP, and ALIGN, respectively, directly demonstrating substantial mitigation of semantic affinity anchoring (SAA). This reduction translates into average HOS gains of 7.7\%, 6.9\%, and 4.3\% across all thresholding strategies (Table~\ref{tab:ablation_suff}), confirming improved known/unknown separability. Figures~\ref{fig:tsne_all} and \ref{fig:heatmap} further illustrate SUFF’s dual effect: it separates unknown classes while preserving known-class separability and markedly reduces their clustering around specific known classes. See Appendix~\ref{appendix:Ablation-Study} for detailed results.

\begin{table}[htbp]
  \centering
     \renewcommand{\arraystretch}{1.2}

    \resizebox{\linewidth}{!}{
    \begin{tabular}{c|c|c|c|c}
    \toprule[1.5pt]
        Method & VTAB-6 & Office-Home & VisDA-2017 & AVG$\downarrow$ \\ \hline
        CLIP & 0.53 & 0.53 & 0.33 & 0.46 \\ 
        CLIP w/ SUFF & \textbf{0.44} & \textbf{0.24} & \textbf{0.00} & \textbf{0.23}{\scriptsize\textbf{($\downarrow$ 0.23)}} \\ \hline
        SigLIP & 0.56 & 0.67 & 0.50 & 0.58 \\ 
        SigLIP w/ SUFF & \textbf{0.32} & \textbf{0.24} & \textbf{0.33} & \textbf{0.30}{\scriptsize\textbf{($\downarrow$ 0.28)}} \\ \hline
        ALIGN & 0.57 & 0.54 & 0.50 & 0.54 \\ 
        ALIGN w/ SUFF & \textbf{0.32} & \textbf{0.14} & \textbf{0.17} & \textbf{0.21}{\scriptsize\textbf{($\downarrow$ 0.33)}} \\ 
        \bottomrule[1.5pt]
    \end{tabular}
}
  \caption{AR@1 on VTAB-6, Office-Home and VisDA-2017 for baseline vs. SUFF-augmented models ($\downarrow$ lower is better.) }
  \label{tab:AR@1}%
\end{table}%

\begin{table}[htbp]
  \centering
  \renewcommand{\arraystretch}{1.2}
  \resizebox{\linewidth}{!}{
    \begin{tabular}{c|c|c|c|c|c}
        \toprule[1.5pt]
    VLM   & w/ SUFF & Fixed & k-means & BGAT & AVG \\ \hline
    \multirow{2}[0]{*}{CLIP} & $\times$    & 66.9  & 74.0    & 76.4 & 72.4  \\
          & $\checkmark$   & \textbf{79.8}{\scriptsize\textbf{(+12.9\%)}}  & \textbf{80.3}{\scriptsize\textbf{(+6.2\%)}}  & \textbf{80.5}{\scriptsize\textbf{(+4.1\%)}} & \textbf{80.1}{\scriptsize\textbf{(+7.7\%)}}\\ \hline
    \multirow{2}[0]{*}{SigLIP} & $\times$    & 52.1  & 69.3  & 80.0 & 67.1 \\
          & $\checkmark$   & \textbf{65.4}{\scriptsize\textbf{(+13.3\%)}}  & \textbf{74.8}{\scriptsize\textbf{(+5.5\%)}} & \textbf{81.4}{\scriptsize\textbf{(+1.4\%)}} & \textbf{73.9}{\scriptsize\textbf{(+6.9\%)}} \\ \hline
    \multirow{2}[0]{*}{ALIGN} & $\times$    & 70.9  & 70.6  & 77.0  & 72.8 \\
          & $\checkmark$   & \textbf{77.4}{\scriptsize\textbf{(+6.5\%)}}  & \textbf{77.2}{\scriptsize\textbf{(+6.6\%)}}  & \textbf{78.5}{\scriptsize\textbf{(+1.5\%)}} & \textbf{77.7}{\scriptsize\textbf{(+4.3\%)}} \\ 
              \bottomrule[1.5pt]
    \end{tabular}%
   }
     \caption{HOS(\%) before vs. after SUFF across different VLM backbones and threshold strategies.}
  \label{tab:ablation_suff}%
\end{table}%

\textbf{Effectiveness of BGAT.}
We compare five thresholding strategies across six tasks using three VLMs: a fixed threshold at half the maximum score (Fixed)~\cite{2025_CVIU_ODAwVL}, the mean of all scores (Mean)~\cite{2022_ECCV_ROS_OSDA}, k-means clustering (K-Means)~\cite{2020_ICML_SHOT}, the intersection point of two Gaussian densities (GMM-Int)~\cite{2025_AAAI_TASC}, and our proposed BGAT method.
As shown in Figure~\ref{fig:threshold_ablation}, BGAT consistently achieves the best performance across all VLMs, both on raw features and SUFF-enhanced features, demonstrating its robustness and adaptability to feature variations. See Appendix~\
ref{appendix:Ablation-Study} for detailed results.

\begin{table}[htbp]
  \centering
  \renewcommand{\arraystretch}{1.1}
    \resizebox{\linewidth}{!}{
    \begin{tabular}{lccccc}
    \toprule[1.5pt]
    \multicolumn{1}{c}{Method} & Art   & Clipart & Product & Real  & AVG \\ \hline
    SHOT  & 58.1  & 50.8  & 59.9  & 62.3  & 57.8 \\
    w/ BGAT & 59.2  & 54.8  & 65.7  & 68.8  & 62.1{\scriptsize\textbf{(+4.3\%)}} \\
    w/ BGAT + SUFF & \textbf{59.3}  & \textbf{55.3}  & \textbf{66.5}  & \textbf{70.5}  & \textbf{62.9}{\scriptsize\textbf{(+5.1\%)}} \\ \hline
    UEO-CLIP & 68.3  & 63.7  & 65.6  & 70.4  & 67.0 \\
    w/ BGAT & 72.1  & 71.1  & 79.5  & 80.7  & 75.9{\scriptsize\textbf{(+8.9\%)}} \\
    w/ BGAT + SUFF & \textbf{74.8}  & \textbf{73.5}  & \textbf{83.6}  & \textbf{83.3}  & \textbf{78.8}{\scriptsize\textbf{(+11.8\%)}} \\
    \bottomrule[1.5pt]
    \end{tabular}%
    }
      \caption{HOS (\%) for SHOT and UEO-CLIP enhanced with our modules on the four Office-Home subdomains.}
  \label{tab:plug_ablation}%
\end{table}%

\begin{figure*}[htbp]
    \centering
    \begin{subfigure}[b]{0.1623\linewidth}
        \includegraphics[width=\textwidth]{figs/TSNE/Art/source_features.pdf}
        \caption{Art}
        \label{fig:tsne_1}
    \end{subfigure}
    \begin{subfigure}[b]{0.16\linewidth}
        \includegraphics[width=\textwidth]{figs/TSNE/Product/source_features.pdf}
        \caption{Product}
        \label{fig:tsne_3}
    \end{subfigure}
    \begin{subfigure}[b]{0.16\linewidth}
        \includegraphics[width=\textwidth]{figs/TSNE/Real_World/source_features.pdf}
        \caption{Real}
        \label{fig:tsne_4}
    \end{subfigure}
    \begin{subfigure}[b]{0.1623\linewidth}
        \includegraphics[width=\textwidth]{figs/TSNE/Art/filterd_features.pdf}
        \caption{Art-S}
        \label{fig:tsne_6}
    \end{subfigure}
    \begin{subfigure}[b]{0.16\linewidth}
        \includegraphics[width=\textwidth]{figs/TSNE/Product/filterd_features.pdf}
        \caption{Product-S}
        \label{fig:tsne_7}
    \end{subfigure}
    \begin{subfigure}[b]{0.16\linewidth}
        \includegraphics[width=\textwidth]{figs/TSNE/Real_World/filterd_features.pdf}
        \caption{Real-S}
        \label{fig:tsne_8}
    \end{subfigure}

    \caption{t-SNE Visualizations of four domains in the Office-Home datasets. (a)-(c) represent the raw features obtained from CLIP for target dataset image samples, while (d)-(f) represent the features filtered by the SUFF module. Blue points indicate unknown class samples, and the points in other colors represent different known classes.}
    \label{fig:tsne_all}
\end{figure*}

\textbf{Plug-and-Play Validation on UEO and SHOT.}
Table~\ref{tab:plug_ablation} demonstrates the effectiveness of our proposed BGAT and SUFF modules when integrated into the VLM-based method UEO and the representative single-modal SF-OSDA method SHOT~\cite{2020_ICML_SHOT}. As shown in Table~\ref{tab:plug_ablation}, adding BGAT improves UEO and SHOT by 8.9\% and 4.3\% in terms of HOS; with SUFF added, these gains rise to 11.8\% and 5.1\%. This confirms the generality of our approach across both multimodal and single-modal methods.

\textbf{Effectiveness of Box-Cox.}
As shown in Figure~\ref{fig:ablation_tau}(a–c), applying the Box–Cox transformation in BGAT consistently improves performance across five tasks and three VLM backbones, confirming its effectiveness in improving threshold estimation and overall accuracy. On average, the three VLMs gain 2.67\%, with SigLIP benefiting the most due to its more concentrated known-class scores, which make Box-Cox more effective. See Appendix~\ref{appendix:Ablation-Study} for more details.

\begin{figure}
    \centering
    \includegraphics[width=\linewidth]{figs/box_and_tau.pdf}
    \caption{Effect of the Box‑Cox transformation and SUFF on HOS (\%) for Office‑Home and VisDA datasets. Panels (a–c) show the impact of  Box‑Cox transformation and panel (d) shows the effect of varying $\tau$.}
    \label{fig:ablation_tau}
\end{figure}

\subsection{Hyper-parameter Analysis}
We analyze the effect of the variance retention ratio ($\tau$) used in subspace construction (Eq.~\ref{equa:svd_tau}).  
Since the singular value spectrum decays rapidly and then flattens, small $\tau$ values may discard useful information, while large $\tau$ values tend to retain noise or redundancy. As a result, extreme values offer no meaningful benefit, and we explore a practical range of [0.7, 0.9]. As shown in Figure~\ref{fig:ablation_tau} (d), the average HOS remains consistently stable across five tasks, with a variance of just 0.19~(\%)$^2$, indicating that model performance is largely insensitive to the choice of $\tau$ within this range.

\subsection{Case Study}
\textbf{Visualization of Semantic Affinity Anchoring.} 
Figure~\ref{fig:heatmap} shows heat maps of the anchoring of unknown to known classes (a) before SUFF module and (b) after SUFF module, with anchored unknowns dropping from 18 to 4. 
This shows SUFF greatly reduces semantic affinity anchoring from unknowns to known classes. The few remaining anchors result from the limits in VLM’s initial sample selection, making it difficult for the unknown‑class subspace to cover the feature distribution of these unknown samples. \textbf{t-SNE Visualization.} Figure~\ref{fig:tsne_all} (a–c) shows the original VLM feature distribution, where unknown-class features are mixed with known-class clusters and hard to distinguish. After applying SUFF (Figure~\ref{fig:tsne_all} (d–f)), unknown features are clearly separated, and the separability among known classes is also preserved. This confirms that SUFF improves known/unknown discrimination without harming known-class structure.

\begin{figure}[]
    \centering
    \begin{minipage}{0.47\linewidth}
        \centering
        \includegraphics[width=\linewidth]{figs/25Custom_Unknown_Tendency.pdf}
        \subcaption{}
        \label{fig:unknow_tendency_clip}
    \end{minipage}%
    \hfill
    \hspace{0.2em}
    \begin{minipage}{0.49\linewidth}
        \centering
        \includegraphics[width=\linewidth]{figs/25Custom_Unknown_Tendency_Real_suff.pdf}
        \subcaption{}
        \label{fig:unknow_tendency_suff}
    \end{minipage} 
\caption{Unknown-class heat map in the Real-World domain of the Office-Home dataset for the ALIGN model: (a) before SUFF processing and (b) after SUFF processing.
}
\label{fig:heatmap}
\end{figure}

\section{Conclusion}
\label{sec:Conclusion}

This work identifies and formally defines \textit{Semantic Affinity Anchoring (SAA)}, a phenomenon common to many benchmark datasets. In SAA, certain unknown-class samples cluster near semantically related known classes in the VLM feature space, which makes them hard to detect. To tackle this problem, we propose VLM-OpenXpert, a completely training-free and label-free inference framework consisting of two plug-and-play modules. Specifically, SUFF filters unknown-class features through soft suppression in a low-rank subspace dominated by unknown-class feature energy, significantly reducing SAA-induced detection failures. BGAT combines a Box–Cox transformation with a bimodal Gaussian mixture model to adaptively estimate an optimal threshold that balances known class recognition and unknown class detection.
With these two lightweight inference modules, VLM‑OpenXpert surpasses or matches training‑intensive state‑of‑the‑art methods on unlabeled target domains, validating its effectiveness and establishing a new design paradigm for unsupervised open‑set domain adaptation of vision-language models.

\section*{Acknowledgements}
This work is supported by the National Natural Science Foundation of China (No.62206107 and No.62406127).

\bibliography{main}

\clearpage

\appendix

\twocolumn[%
  \begin{center}
    {\Large \bfseries Appendix}
  \end{center}
  \vspace{1em}
]
\section{Proof}
\label{appendix:Proof}

\begin{table*}[]
    \renewcommand{\arraystretch}{1.8}
    \centering
    \resizebox{\textwidth}{!}{
        \begin{tabular}{c|c|c|c|c|c|c|c|c|c}
            \toprule[1.5pt]
            \textbf{Setting} & \textbf{ImageNet}  & \textbf{DTD} & \textbf{EuroSAT} & \textbf{Food101}  & \textbf{OxfordPets}  & \textbf{StanfordCars} &  \textbf{Office-Home} & \textbf{VisDA-2017} & \textbf{DomainNet} \\ 
            \midrule
            $C_{know}$ & 200   & 20 & 5 & 35  & 20  & 80  & 25 & 6 & 100 \\ 
            \midrule
            $C_{unk}$ & 800  & 17 & 5 & 66  & 17  & 116  & 40 & 6 & 245 \\ 
            \bottomrule[1.5pt]
        \end{tabular}
    }
        \caption{Setting of the Number of Known and Unknown Classes in Different Datasets}
        \label{table:dataset_settings}  
\end{table*}

\textbf{Theorem 1:}

Let the known class follow a Gaussian distribution $N(\mu_1, \sigma_1^2)$ and the unknown class follow a Gaussian distribution $N(\mu_2, \sigma_2^2)$, where $\mu_1 < \mu_2$. Define the total classification error area as:
$$
E(t) = \int_{-\infty}^{t} f_2(x)dx + \int_{t}^{\infty} f_1(x)dx
$$
where $f_1(x)$ and $f_2(x)$ denote the probability density functions (PDFs) of the two distributions. Then, the optimal threshold $T^*$ that minimizes $E(t)$ satisfies:
$$
 f_1(T^*) = f_2(T^*)
$$
i.e., $T^*$ is the x-coordinate of the intersection point of the two Gaussian PDFs.

\textbf{Proof.} To find the threshold $T^*$ that minimizes the total classification error $E(t)$, we analyze the structure of the error function as follows:

\textbf{1. Construction of the Error Function  }

   The total classification error $E(t)$ consists of two components:

   (1) The first term $\int_{-\infty}^{t} f_2(x)dx$: it denotes the probability that an unknown sample is misclassified as a known sample (false negative)
   
   (2) The second term $\int_{t}^{\infty} f_1(x)dx$: it denotes the probability that a known sample is misclassified as an unknown sample (false positive).

\textbf{2. Condition for Extremum}

   Taking the derivative of $E(t)$ with respect to $t$, we obtain:
   $$
   \frac{dE}{dt} = f_2(t) - f_1(t)
   $$  
   Setting the derivative to zero to find the critical point:
   $$
   f_1(T^*) = f_2(T^*)
   $$

\textbf{3. Second Derivative Test}

   Computing the second derivative:
   $$
   \frac{d^2E}{dt^2} = f_2'(t) - f_1'(t)
   $$  
   At the intersection point $T^*$:
   when $\mu_1 < T^* < \mu_2$, we have $f_1'(T^*) > 0$ (since $f_1$ is increasing on the right) and $f_2'(T^*) < 0$ (since $f_2$ is decreasing on the right).
   Consequently, $\frac{d^2E}{dt^2} \big|_{t=T^*} > 0$, confirming that $T^*$ is a minimum.

\textbf{Conclusion :} Under the assumption of bimodal Gaussian distributions, the optimal threshold that minimizes the total classification error corresponds to the x-coordinate of the intersection of the two probability density functions. 

\textbf{Theorem 2:} 

Let the known class follow  $N(\mu_1, \sigma_1^2)$  and the unknown class follow $ N(\mu_2, \sigma_2^2)$, with probability density functions (PDFs) defined as:  
\[
f_1(x) = \frac{1}{\sigma_1\sqrt{2\pi}} e^{-\frac{(x-\mu_1)^2}{2\sigma_1^2}}, \quad f_2(x) = \frac{1}{\sigma_2\sqrt{2\pi}} e^{-\frac{(x-\mu_2)^2}{2\sigma_2^2}}.
\]

\textbf{Step 1: Intersection Equation}

The intersection of the two distributions satisfies $ f_1(x) = f_2(x)$. Simplifying this equality yields:  
\[
\frac{(x-\mu_1)^2}{\sigma_1^2} - \frac{(x-\mu_2)^2}{\sigma_2^2} = 2\ln\left(\frac{\sigma_2}{\sigma_1}\right). \tag{1}
\]

\textbf{Step 2: Implicit Function Differentiation}

Treating Equation (1) as an implicit function $ F(x, \sigma_1) = 0$, we apply the implicit function theorem to compute $ \frac{dx}{d\sigma_1}$:  
\[
\frac{dx}{d\sigma_1} = -\frac{\partial F/\partial \sigma_1}{\partial F/\partial x},
\]  
where the partial derivatives are:  
\[
\frac{\partial F}{\partial \sigma_1} = -\frac{2(x-\mu_1)^2}{\sigma_1^3} + \frac{2}{\sigma_1}, \quad \frac{\partial F}{\partial x} = \frac{2(x-\mu_1)}{\sigma_1^2} - \frac{2(x-\mu_2)}{\sigma_2^2}.
\]

\textbf{Step 3: Sensitivity Analysis} 

As $ \sigma_1 \to 0$:  
(1) The dominant term in $ \partial F/\partial \sigma_1$ is $ -\frac{2(x-\mu_1)^2}{\sigma_1^3}$.  

(2) The dominant term in $ \partial F/\partial x$ is $ \frac{2(x-\mu_1)}{\sigma_1^2}$.  

Thus,  
\[
\frac{dx}{d\sigma_1} \approx \frac{\frac{2(x-\mu_1)^2}{\sigma_1^3}}{\frac{2(x-\mu_1)}{\sigma_1^2}} = \frac{x-\mu_1}{\sigma_1}.
\]

\textbf{Conclusion :} When $ \sigma_1$ is extremely small, the magnitude of $ \frac{dx}{d\sigma_1}$ scales as $ \frac{x-\mu_1}{\sigma_1}$. Even if $ x$ is near $ \mu_1$ (i.e., the intersection point $ T^*$ is close to the known class mean), this derivative remains significant. This demonstrates that the intersection coordinate $ T^*$ is highly sensitive to infinitesimal changes in $ \sigma_1$. The sensitivity arises from the amplification effect caused by the cubic ($ \sigma_1^3$) and quadratic ($ \sigma_1^2$) terms of $ \sigma_1$ in the denominators of the partial derivatives.

\section{Experiments}
\label{appendix:Experiments}


\subsection{Datasets}
\label{appendix:Experiments-Datasets}

\begin{table*}[htbp]
  \centering
  \renewcommand{\arraystretch}{1.2}
     \resizebox{\linewidth}{!}{
    \begin{tabular}{c|cccccc|cccccc|c}
    \toprule[1.5pt]
    \multirow{2}[0]{*}{Method} & \multicolumn{6}{c|}{VATB-6}                    & \multicolumn{6}{c|}{DomainNet}                 &  \\ \cline{2-14}
          & ImageNet & DTD   & EuroSAT & Food101 & OxfordPets & StandfordCars & \multicolumn{1}{c}{C} & \multicolumn{1}{c}{I} & \multicolumn{1}{c}{P} & \multicolumn{1}{c}{Q} & \multicolumn{1}{c}{R} & \multicolumn{1}{c|}{S} & \multicolumn{1}{c}{AVG} \\ \cline{2-14}
    MCM-CLIP & 61.5  & 52.7  & 51.1  & 74.6  & 68.2  & 51.4  & 75.4  & 66.3  & 73.5  & 38.4  & 79.2  & 74.2  & 63.9 \\
    MCM-SigLIP & 63.5  & 68.2  & 50.2  & 77.9  & 75.4  & 62.6  & 75.3  & 67.6  & 72.4  & 33    & 77.2  & 73.5  & 66.4 \\
    MCM-ALIGN & 57.7  & 62.9  & 40.8  & 73.3  & 63.6  & 57.2  & 76.2  & 68.7  & 72.2  & 28.4  & 80.4  & 73.3  & 62.9 \\
    UEO-CLIP & 56.7  & 51    & 44.1  & 70.1  & 69.4  & 50.5  & 77.4  & 69.5  & 75    & 39.1  & 81.5  & 74.9  & 63.3 \\
    UOTA-CLIP & -     & -     & -     & -     & -     & -     & 82.4  & 68.7  & 77    & 35.9  & 85.2  & 77.3  & \multicolumn{1}{c}{-} \\
    our-CLIP & 71.5 $\pm$ 0.0  & 53.2 $\pm$ 0.0  & \textbf{51.9} $\pm$ 0.0 & \textbf{86.6} $\pm$ 0.0 & 73.4 $\pm$ 0.0  & 55.7 $\pm$ 0.0  & \textbf{82.9 $\pm$ 0.0}  & \textbf{69.1 $\pm$ 0.0} & \textbf{79.0 $\pm$ 0.0} & \textbf{39.4 $\pm$ 0.0} & \textbf{87.3 $\pm$ 0.0} & \textbf{79.8 $\pm$ 0.0} & 69.1 $\pm$ 0.0 \\
    our-SigLIP & \textbf{69.9 $\pm$ 0.0} & 64.6 $\pm$ 0.0  & 52.3 $\pm$ 0.0  & 84.1 $\pm$ 0.0  & \textbf{79.7 $\pm$ 0.0} & \textbf{70.7 $\pm$ 0.0} & 79.8 $\pm$ 0.0  & 68.9 $\pm$ 0.0  & 77.2 $\pm$ 0.0  & 35.8 $\pm$ 0.0  & 84.1 $\pm$ 0.0  & 78.8 $\pm$ 0.0  & \textbf{70.5 $\pm$ 0.0} \\
    our-ALIGN & 60.7 $\pm$ 0.0  & \textbf{68.3 $\pm$ 0.0} & 39.1 $\pm$ 0.0  & 79.6 $\pm$ 0.0  & 66.3 $\pm$ 0.0  & 59.1 $\pm$ 0.0  & 76.6 $\pm$ 0.0  & 68.2 $\pm$ 0.0  & 73.1 $\pm$ 0.0  & 29.1 $\pm$ 0.0  & 82.5 $\pm$ 0.0  & 71.7 $\pm$ 0.0  & 64.5 $\pm$ 0.0 \\
    \bottomrule[1.5pt]
    \end{tabular}%
    }
      \caption{HOS (\%) on VATB and DomainNet (mean $\pm$ std) across different methods.
}
  \label{tab:detailed_vatb_and_domainnet}%
\end{table*}%

To comprehensively evaluate the effectiveness of the proposed method, we conduct extensive experimental validation on multiple public datasets. Given that our method does not require source domain data, we first test it on four widely used open-set domain adaptation (OSDA) benchmark datasets: Office-Home~\cite{2017_CVPR_OfficeHome} (65 categories, 4 domains), VisDA-2017~\cite{2017_arXiv_VisDA} (12 categories, cross-domain adaptation from synthetic to real-world scenes), DomainNet~\cite{2019_ICCV_DomainNet} (345 categories, 6 domains). Furthermore, to further assess the generalization ability of our method, we select 6 publicly available datasets from the Visual Task Adaptation Benchmark (VTAB) for experimentation, including ImageNet~\cite{2009_CVPR_ImageNet},  EuroSAT~\cite{2019_J-STAR_EuroSAT}, Food101~\cite{2014_ECCV_Food101},  OxfordPets~\cite{2012_CVPR_Oxfordpets},  StandfordCars~\cite{2013_ICCVW_OxfordCars} and DTD~\cite{2014_CVPR_DTD}. These datasets cover diverse visual tasks, ranging from fine-grained classification to scene recognition, with characteristics such as large category numbers (100-1000 categories), large data scales (tens of thousands to millions of samples), and significant domain differences, providing a comprehensive testing platform for evaluating open-set domain adaptation methods.

\begin{table*}[]
    \centering
    \renewcommand{\arraystretch}{1.2}
    \resizebox{\linewidth}{!}{
    \begin{tabular}{c|c|c|c|c|c|c|c|c|c|c|c|c}
    \toprule[1.5pt]
        Method & Art & Clipart & Product & Real & Visda & ImageNet & DTD & EuroSAT & Food101 & OxfordPets & StandfordCars & $\downarrow$ AVG \\ \hline
        CLIP & 0.40 & 0.55 & 0.65 & 0.53 & 0.33 & 0.39 & 0.29 & 0.80 & 0.32 & 0.76 & 0.62 & 0.51 \\ 
        CLIP w/ SUFF & \textbf{0.15} & \textbf{0.35} & \textbf{0.28} & \textbf{0.17} & \textbf{0.00} & \textbf{0.24} & \textbf{0.24} & \textbf{0.80} & \textbf{0.15} & \textbf{0.65} & \textbf{0.58} & \textbf{0.33 ($\downarrow$ 0.18)}  \\ \hline
        Siglip & 0.55 & 0.82 & 0.75 & 0.57 & 0.50 & 0.46 & 0.53 & 0.40 & 0.36 & 0.76 & 0.84 & 0.59 \\ 
        Siglip w/ SUFF & \textbf{0.20} & \textbf{0.38} & \textbf{0.17} & \textbf{0.20} & \textbf{0.33} & \textbf{0.23} & \textbf{0.12} & \textbf{0.60} & \textbf{0.08} & \textbf{0.35} & \textbf{0.53} & \textbf{0.29 ($\downarrow$ 0.30)}\\ \hline
        ALIGN & 0.42 & 0.65 & 0.60 & 0.47 & 0.50 & 0.41 & 0.41 & 0.80 & 0.33 & 0.76 & 0.72 & 0.55 \\ 
        ALIGN w/ SUFF & \textbf{0.12} & \textbf{0.15} & \textbf{0.15} & \textbf{0.12} & \textbf{0.17} & \textbf{0.26} & \textbf{0.06} & \textbf{0.40} & \textbf{0.15} & \textbf{0.53} & \textbf{0.54} & \textbf{0.24 ($\downarrow$ 0.31)}\\ \bottomrule[1.5pt]
    \end{tabular}
    }
    \caption{AR@1 before and after applying the SUFF module on VATB-6, Office-Home (Art, Clipart, Product, Real), and VisDA-2017 datasets.}
    \label{tab:detailed_AR1}
\end{table*}


In the experiments, the selection criteria for known classes in the four OSDA datasets followed the UOTA~\cite{2023_ICMLW_UOTA} setting to ensure comparability and consistency. For the remaining 6 VATB datasets, we test with varying proportions of unknown classes to enhance the generality and robustness of the experiments, with the specific division of known and unknown classes for each dataset shown in Table \ref{table:dataset_settings}. Specifically, we sort the class names of each dataset in dictionary order, assigning the first $C_{\text{know}}$ classes as known classes, while the rest are considered unknown classes. This diversified testing setup not only validates the adaptability of the method across different data distributions but also comprehensively evaluates its performance on open-set domain adaptation tasks.
Additionally, it should be noted that for the 6 VTAB datasets, we directly conduct testing on the test set and evaluate performance. The test set partitioning follows the approach of CoOp~\cite{IJCV_2022_CoOp}.

\textbf{Implementation Details.}
 All results are averaged over 5 independent runs with random seeds 1, 2, 3, 4 and 5. Since our method involves inference only and requires no training, the variance is negligible (effectively near zero). Experiments were performed on an Intel Xeon Gold 6442Y CPU, a Tesla L40 GPU, and 512 GB of RAM running Ubuntu 18.04. The software environment consists of Python 3.8.20 and PyTorch 2.0.1. Unless stated otherwise, we used a batch size of 32 and a temperature $\tau = 0.8 $. Since our method is training-free, no learning-rate hyperparameter is required.

\subsection{Baselines}
\label{appendix:Experiments-Baselines}

To the best of our knowledge, the proposed Source-Free open-set domain adaptation (SF-OSDA) method, which requires neither source domain data nor training, has not been explored in existing research. Therefore, we select the following three categories of baseline methods for comparison to ensure the comprehensiveness and fairness of the evaluation:

\begin{itemize} 

\item  The VLM-based zero‑shot out‑of‑distribution detection method MCM~\cite{2022_NIPS_MCM}. In our experiments, as a baseline, we adopt the commonly used and relatively stable approach in OSDA by using the mean score of all test samples as the default threshold, since the original paper does not specify a concrete thresholding strategy.

\item Unsupervised open‑set domain adaptation (OSDA) methods jointly train on labeled source and unlabeled target domains, including USD~\cite{2024_CVPR_USD}, UniOT~\cite{NIPS_2022_UniOT}, OSMDA~\cite{TCSVT_2025_OSMDA} and GLC~\cite{CVPR_2023_GLC}. During training, these methods treat all labeled source-domain classes as known and regard any target-domain classes outside this set as unknown.

\item Source‑free unsupervised OSDA methods use a model trained on the labeled source domain as a reference for unsupervised training on the unlabeled target domain, including LAED~\cite{2024_CVPR_LEAD}, DIFO~\cite{CVPR_2024_DIFO}, Co-learn++~\cite{IJCV_2025_COLEARN},COSDA~\cite{2025_ICML_COSDA}, UPUK~\cite{CVPR_2024_UPUK}, COCA\cite{ACCV_2024_COCA}, COSDA~\cite{2025_ICML_COSDA} and DIFO~\cite{CVPR_2024_DIFO}. 
These methods designate the classes encountered during source-domain training as known and treat all other target-domain classes as unknown.

\item Target-only unsupervised open-set domain adaptation methods, such as UOTA~\cite{2023_ICMLW_UOTA} and UEO~\cite{2024_UEO_ICML}, perform training exclusively on the unlabeled target-domain data. They define a predefined set of known classes and treat all other classes in the target data as unknown.

\end{itemize}

Through the comparison of these three categories of baseline methods, we can comprehensively assess the performance of the proposed method in different scenarios and validate its unique advantages in the absence of source domain data and training conditions.

Additionally,  to facilitate performance comparison with other OSDA methods, we use the commonly adopted HOS metric, which is the harmonic mean of the average accuracy for known classes and the accuracy for unknown classes. The calculation formula is as follows:

\begin{equation}
    HOS = \frac{2 \cdot \text{Acc}_{\text{known}} \cdot \text{Acc}_{\text{unknown}}}{\text{Acc}_{\text{known}} + \text{Acc}_{\text{unknown}}},
\end{equation}
where $\text{Acc}_{\text{known}}$ represents the average accuracy of known classes, and  $\text{Acc}_{\text{unknown}}$ denotes the accuracy of unknown class recognition. This metric reflects the overall performance of the model on both known and unknown classes, balancing the recognition ability for known classes and the detection ability for unknown classes. Therefore, HOS provides an effective standard for evaluating OSDA methods.

\subsection{Overall Performance}
\label{appendix:overall_performance}
In this section, we present the detailed data for the results shown in Table~\ref{tab:vatb_domainnet_performance} of the main text. To ensure a fair comparison, both UEO and MCM use the same threshold‐selection method in Table~\ref{tab:vatb_domainnet_performance}—namely, the average of all sample scores, which is a common and stable choice. The full results are listed in Table~\ref{tab:detailed_vatb_and_domainnet}.

From the detailed results in Table~\ref{tab:detailed_vatb_and_domainnet}, our method delivers significant improvements over the VLM‑based MCM on every dataset. In particular, on ImageNet under CLIP, we see up to a 10 \% gain, and even on DTD—where CLIP’s zero‑shot performance is relatively weak—we achieve a 0.5 \% uplift. This demonstrates our approach’s robustness to noise in VLM predictions, maintaining stability even when zero‑shot accuracy is low. Moreover, on the most challenging domain‑adaptation dataset, DomainNet, we outperform the previous training‑intensive SOTA method UOTA across all domains. These findings confirm the effectiveness of our lightweight inference strategy for VLMs in OSDA scenarios, achieving SOTA performance without any additional training.

\textbf{Performance Comparison: VLM‑OpenXpert vs. ChatGPT‑4o and 4o‑mini}

To evaluate the ability of our method to reject unknown classes compared to the latest cross-modal large language models, we conducted experiments on three datasets of varying scales: the Real domain of OfficeHome, the Clipart domain of DomainNet, and Flowers102~\cite{2009_ICCV_Flowers102}. We crafted refined prompts to guide ChatGPT‑4o and 4o‑mini in determining whether an image belongs to a given category set—if so, they classify it. Their performance is then compared with our lightweight model, VLM-OpenXpert.

\begin{table}[htbp]
  \centering
  \renewcommand{\arraystretch}{1.2}
  \resizebox{\linewidth}{!}{
    \begin{tabular}{c|c|c|c|c}
    \toprule[1.5pt]
    Method & Officehome-R & DomainNet-C & Flowers102 & AVG \\  \hline
    ChatGPT-4o & 89.2  & 55.0  & 53.5  & 65.9  \\
    ChatGPT-4o-mini & 84.8  & 55.0  & 52.7  & 64.1  \\
    our-CLIP (Vit-L/14) & \textbf{89.5 $\pm$ 0.0} & \textbf{83.2 $\pm$ 0.0}  & \textbf{70.2 $\pm$ 0.0}  & \textbf{81.0 $\pm$ 0.0}  \\
    our-CLIP (Vit-B/16) & 85.4 $\pm$ 0.0  & 75.6 $\pm$ 0.0  & 68.8 $\pm$ 0.0  & 76.6 $\pm$ 0.0  \\
    \bottomrule[1.5pt]
    \end{tabular}%
    }
      \caption{HOS (\%) Performance Comparison with Cross‑Modal LLMs (ChatGPT‑4o and 4o‑mini) }
  \label{tab:comparision_with_gpt}%
\end{table}%

We provided ChatGPT with the prompt: "Given the image, determine if the main object belongs to one of these categories: {CLASS NAMES}. If yes, reply with the category name. Otherwise, reply ``unknown". Here, {CLASS NAMES} is replaced with the list of known categories from the dataset.
Based on its responses, we performed text matching to evaluate its ability to reject unknown classes and recognize known ones. As shown in Table~\ref{tab:comparision_with_gpt}, on the small-scale Office-Home Real domain (with only 25 known classes), ChatGPT‑4o achieved performance comparable to our ViT-L/14-based model. However, as the dataset scale and number of categories increased—for example, DomainNet-C (100 classes) and Flowers102 (65 classes)—ChatGPT's performance dropped sharply and fell far behind our method. This decline is mainly due to ChatGPT being designed as a general-purpose vision-language model, not specifically for classification tasks. These results highlight that despite the impressive capabilities of large cross-modal models, developing lightweight VLMs remains highly valuable for fundamental tasks like image classification.

\subsection{Ablation Study}
\label{appendix:Ablation-Study}

This section provides additional ablation study experiments that are not detailed in the main text, including the exact numerical values of some tables and figures, as well as ablation experiments omitted due to space constraints. A more comprehensive dataset and ablation results are presented in this subsection.


\begin{table}[htbp]
  \centering
    \renewcommand{\arraystretch}{1.2}

    \resizebox{\linewidth}{!}{
    \begin{tabular}{c|c|c|cccc|c|c}
    \toprule[1.5pt]
    \multirow{2}[0]{*}{Method} & \multirow{2}[0]{*}{w/ SUFF} & \multirow{2}[0]{*}{ImageNet} & \multicolumn{4}{c|}{OfficeHome} & \multirow{2}[0]{*}{VisDA} & \multirow{2}[0]{*}{AVG} \\ \cline{4-7} 
          &       &       & Art   & Clipart & Product & Real  &       &  \\ \hline
    \multirow{2}[0]{*}{our-CLIP} & $\times$    & 61.6  & 61.9  & 59.9  & 46.4  & 52.8  & 83.8  & 61.1  \\
          & $\checkmark$   & \textbf{69.6}  & \textbf{80.2}  & \textbf{72.1}  & \textbf{85.7}  & \textbf{85.0}  & \textbf{89.0}  & \textbf{80.3}  \\ \hline
    \multirow{2}[0]{*}{our-SigLIP} & $\times$    & 52.6  & 53.0  & 42.8  & 48.7  & 47.4  & 55.6  & 50.0  \\
          & $\checkmark$   & \textbf{64.1 } & \textbf{67.7}  & \textbf{47.8}  & \textbf{68.8}  & \textbf{66.4}  & \textbf{69.5}  & \textbf{64.0}  \\ \hline
    \multirow{2}[0]{*}{ALIGN} & $\times$    & 57.1  & 78.4  & 73.1  & 89.9  & 79.9  & 75.1  & 75.6  \\
          & $\checkmark$   & \textbf{59.6}  & \textbf{84.4}  & \textbf{80.7}  & \textbf{90.5}  & \textbf{90.3}  & \textbf{86.1}  & \textbf{81.9}  \\
          \bottomrule[1.5pt]
    \end{tabular}%
    }
      \caption{Impact of the SUFF module on HOS (\%) for different VLM models under a fixed threshold function.
}
  \label{tab:appendix_detail_suff_fixed}%
\end{table}%

\begin{table}[htbp]
  \centering
      \renewcommand{\arraystretch}{1.2}
  \resizebox{\linewidth}{!}{
    \begin{tabular}{c|c|c|cccc|c|c}
    \toprule[1.5pt]
    \multirow{2}[0]{*}{Method} & \multirow{2}[0]{*}{w/ SUFF} & \multirow{2}[0]{*}{ImageNet} & \multicolumn{4}{c|}{OfficeHome} & \multirow{2}[0]{*}{VisDA} & \multirow{2}[0]{*}{AVG} \\ \cline{4-7} 
          &       &       & Art   & Clipart & Product & Real  &       &  \\ \hline
    \multirow{2}[0]{*}{our-CLIP} & $\times$    & 62.1  & 74.7  & 69.3  & 72.7  & 77.4  & 86.3  & 73.7  \\
          & $\checkmark$   & \textbf{70.8}  & \textbf{79.9}  & \textbf{73.1}  & \textbf{86.2}  & \textbf{85.2}  & \textbf{89.0}  & \textbf{80.7}  \\ \hline
    \multirow{2}[0]{*}{our-SigLIP} & $\times$    & 60.9  & 76.5  & 65.0  & 70.1  & 73.4  & 75.8  & 70.3  \\
          & $\checkmark$   & \textbf{66.5}  & \textbf{80.6}  & \textbf{67.8}  & \textbf{80.8}  & \textbf{81.3}  & \textbf{80.3}  & \textbf{76.2}  \\ \hline
    \multirow{2}[0]{*}{ALIGN} & $\times$    & 52.6  & 81.7  & 77.0  & 90.6  & 84.7  & 75.8  & 77.1  \\
          & $\checkmark$   & \textbf{64.1}  & \textbf{84.5}  & \textbf{82.2}  & \textbf{90.9}  & \textbf{90.9}  & \textbf{80.3}  & \textbf{82.1}  \\
          \bottomrule[1.5pt]
    \end{tabular}%
    }
      \caption{Impact of SUFF on HOS (\%) for different VLM models using a K‑means threshold function.}
  \label{tab:appendix_detail_suff_kmeans}%
\end{table}%


\begin{table}[htbp]
  \centering
  \renewcommand{\arraystretch}{1.2}
  \resizebox{\linewidth}{!}{
    \begin{tabular}{c|c|c|cccc|c|c}
    \toprule[1.5pt]
    \multirow{2}[0]{*}{Method} & \multirow{2}[0]{*}{w/ SUFF} & \multirow{2}[0]{*}{ImageNet} & \multicolumn{4}{c|}{OfficeHome} & \multirow{2}[0]{*}{VisDA} & \multirow{2}[0]{*}{AVG} \\ \cline{4-7}
          &       &       & Art   & Clipart & Product & Real  &       &  \\ \hline
    \multirow{2}[0]{*}{our-CLIP} & $\times$    & 63.3  & 76.5  & 71.7  & 82.1  & 82.8  & 87.8  & 77.3  \\
          & $\checkmark$   & \textbf{71.3}  & \textbf{78.2}  & \textbf{74.0}  & \textbf{86.8}  & \textbf{85.7}  & \textbf{89.0}  & \textbf{80.8}  \\ \hline
    \multirow{2}[0]{*}{our-SigLIP} & $\times$    & 67.2  & 83.7  & \textbf{80.8}  & 89.2  & 88.2  & 87.4  & 82.7  \\
          & $\checkmark$   & \textbf{69.9}  & \textbf{85.8}  & 80.4  & \textbf{89.4}  & \textbf{88.9}  & \textbf{88.1}  & \textbf{83.7}  \\ \hline
    \multirow{2}[0]{*}{ALIGN} & $\times$    & 59.4  & 81.4  & 81.1  & 90.8  & 88.9  & 86.1  & 81.3  \\
          & $\checkmark$   & \textbf{60.7}  & \textbf{85.0}  & \textbf{83.2}  & \textbf{90.9}  & \textbf{91.3}  & \textbf{87.3}  & \textbf{83.1}  \\
           \bottomrule[1.5pt]
    \end{tabular}%
    }
      \caption{Impact of SUFF on HOS(\%) for different VLM models using our BGAT threshold function. }
  \label{tab:appendix_detail_suff_bgat}%
\end{table}%


\begin{figure*}[ht]
    \centering
    \subfloat[Art (CLIP)]{\includegraphics[width=0.23\textwidth]{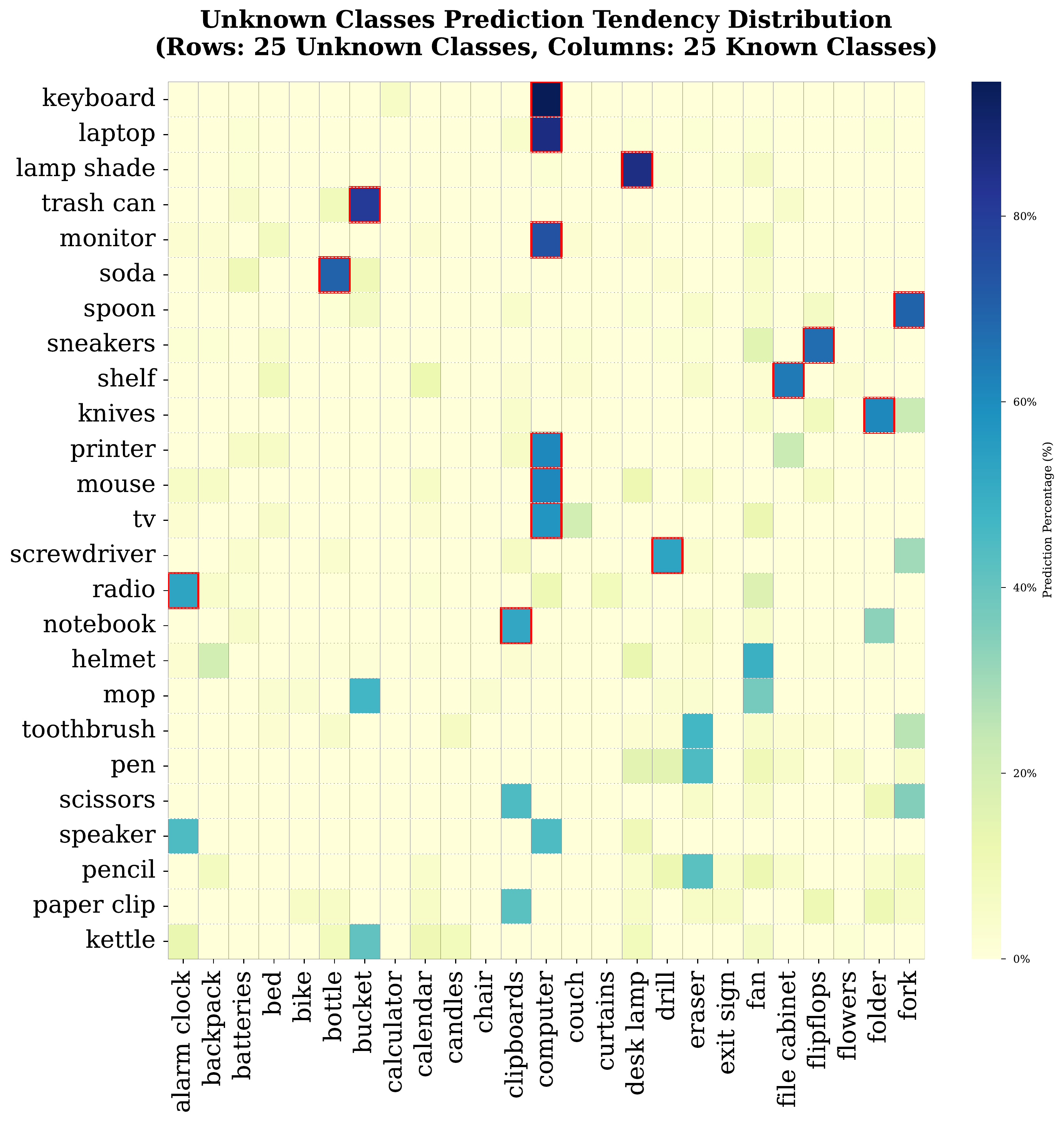}}
    \hfill
    \subfloat[Clipart (CLIP)]{\includegraphics[width=0.23\textwidth]{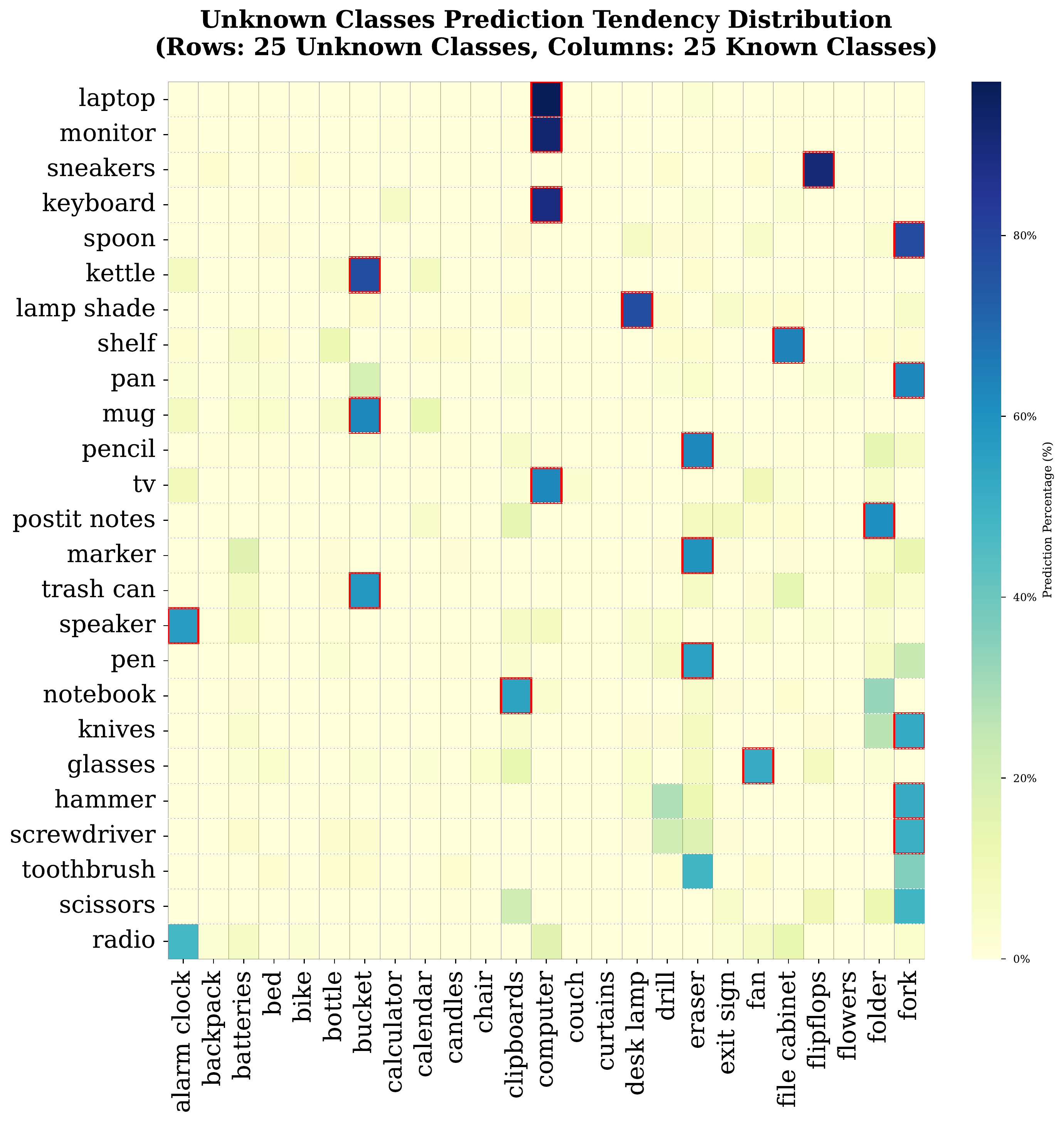}}
    \hfill
    \subfloat[Product (CLIP)]{\includegraphics[width=0.23\textwidth]{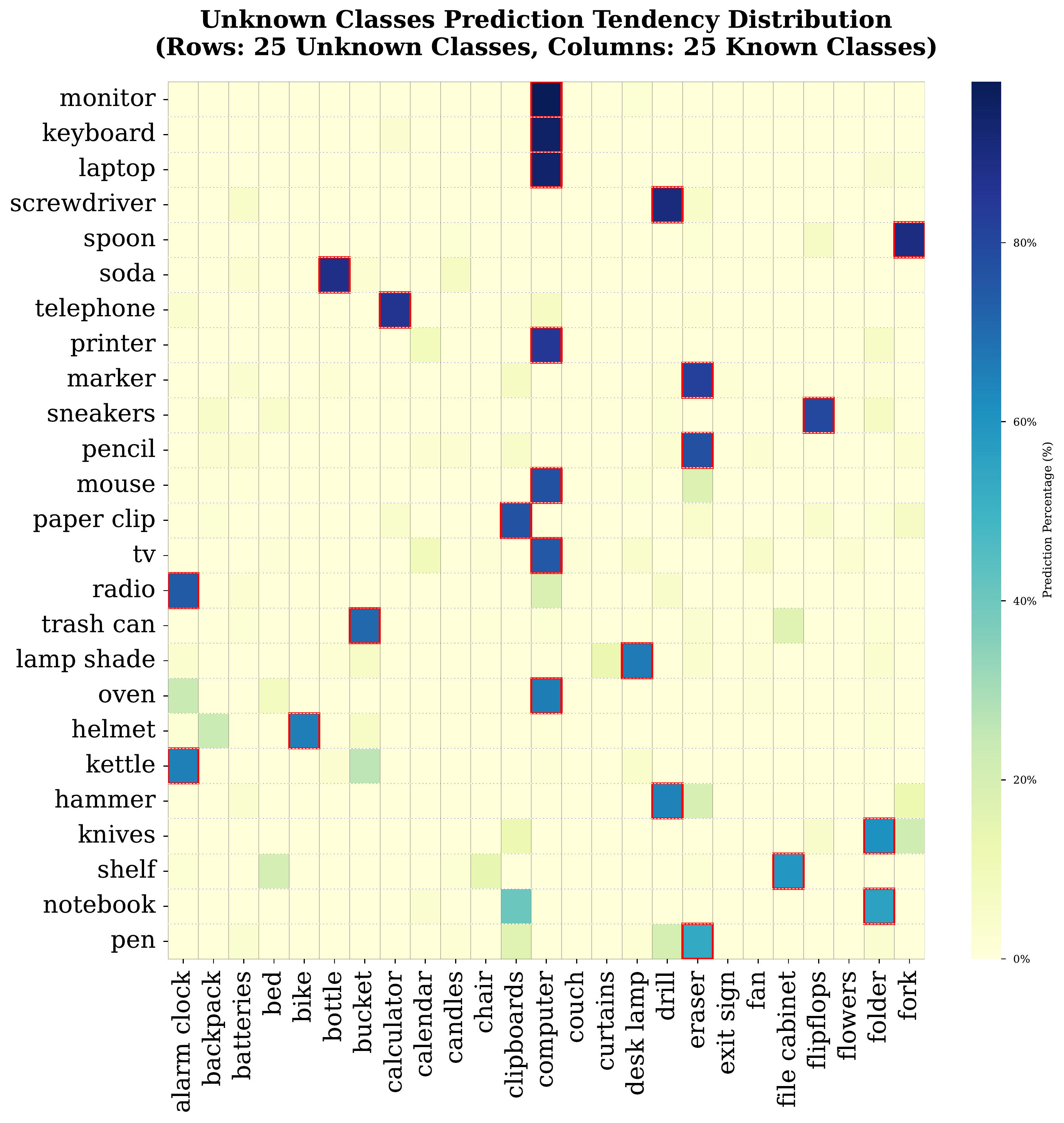}}
    \hfill
    \subfloat[Real World (CLIP)]{\includegraphics[width=0.23\textwidth]{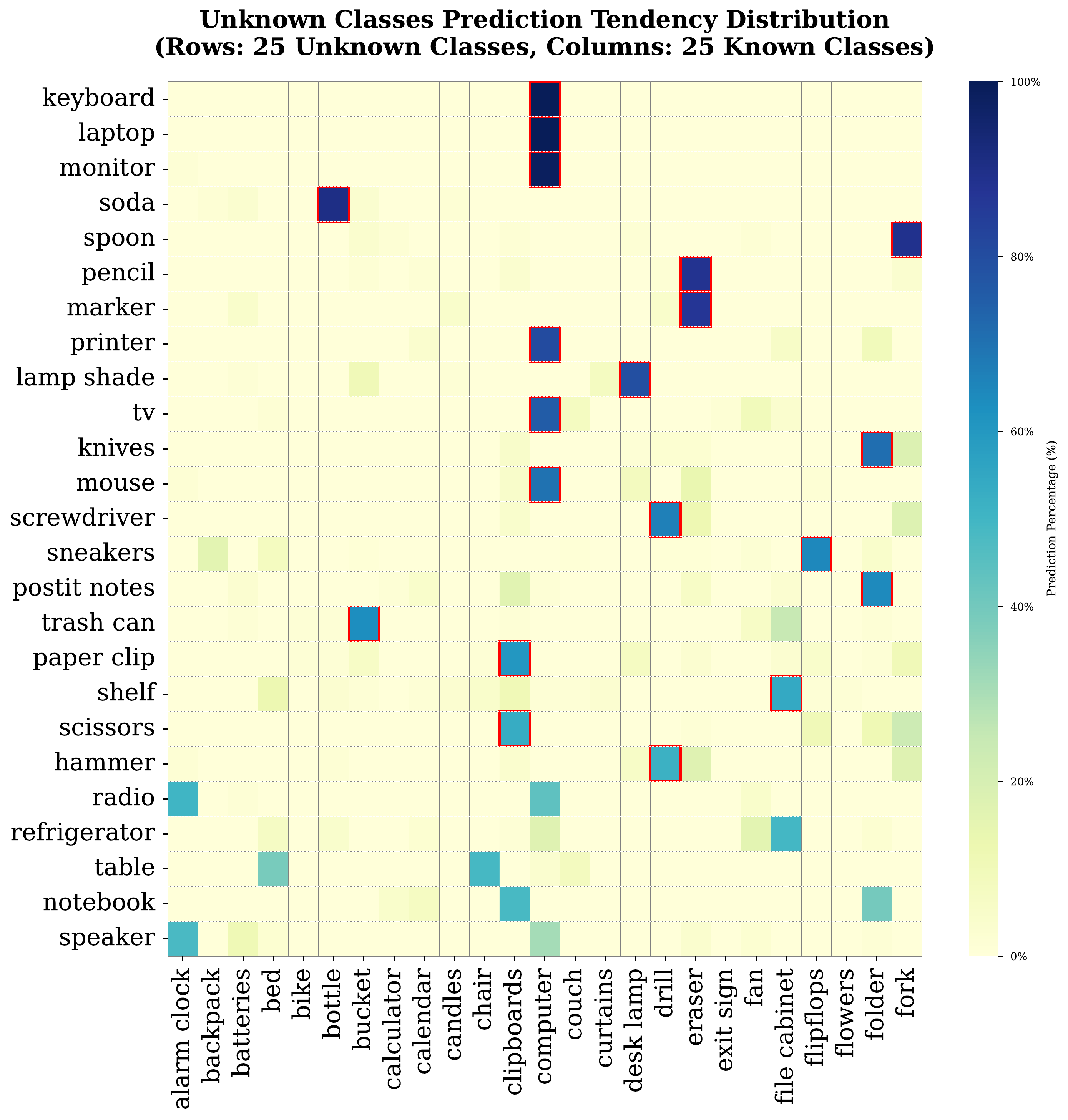}}
    
    \vspace{5pt}
    
    \subfloat[Art (CLIP + SUFF)]{\includegraphics[width=0.23\textwidth]{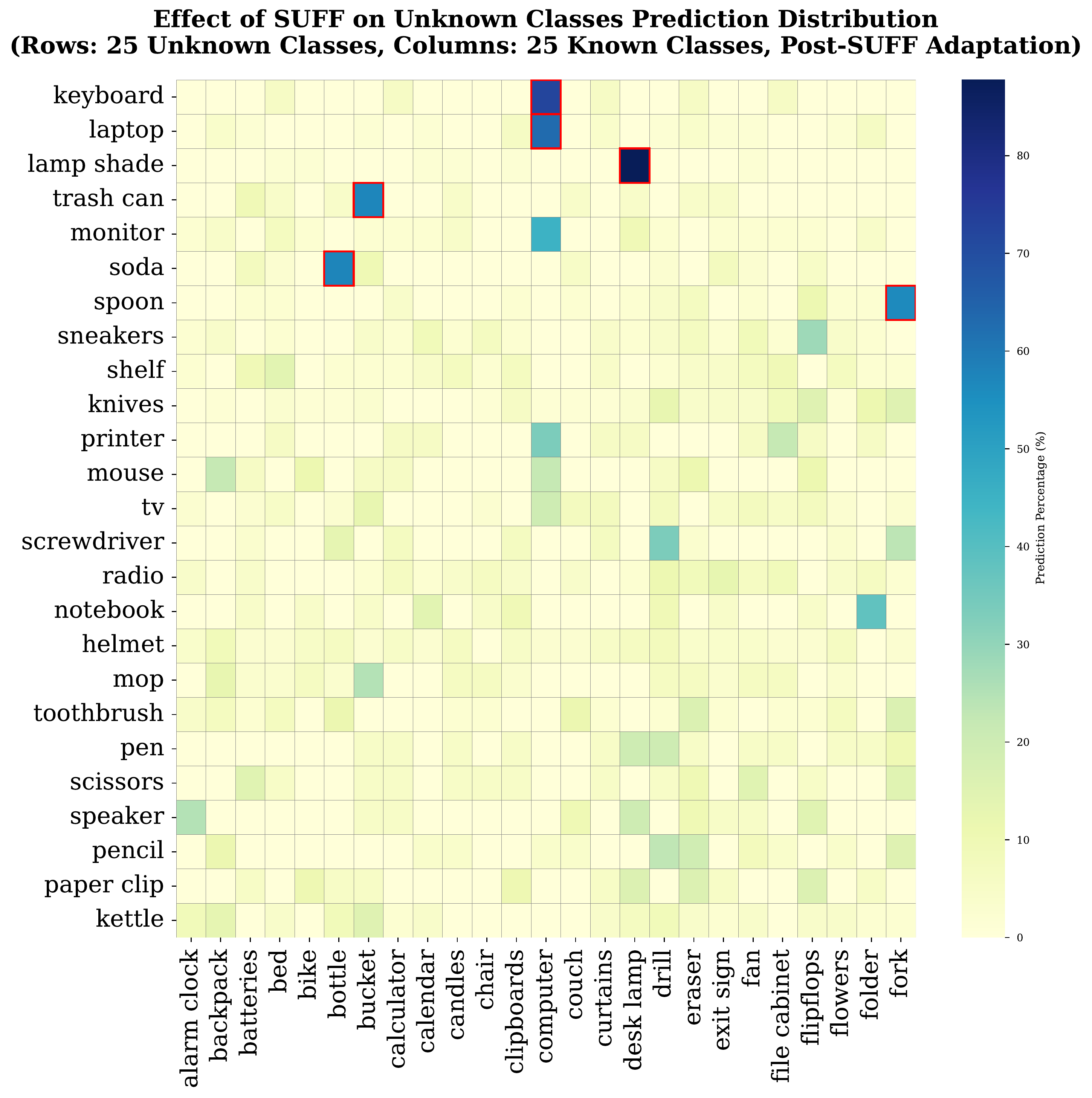}}
    \hfill
    \subfloat[Clipart (CLIP + SUFF)]{\includegraphics[width=0.23\textwidth]{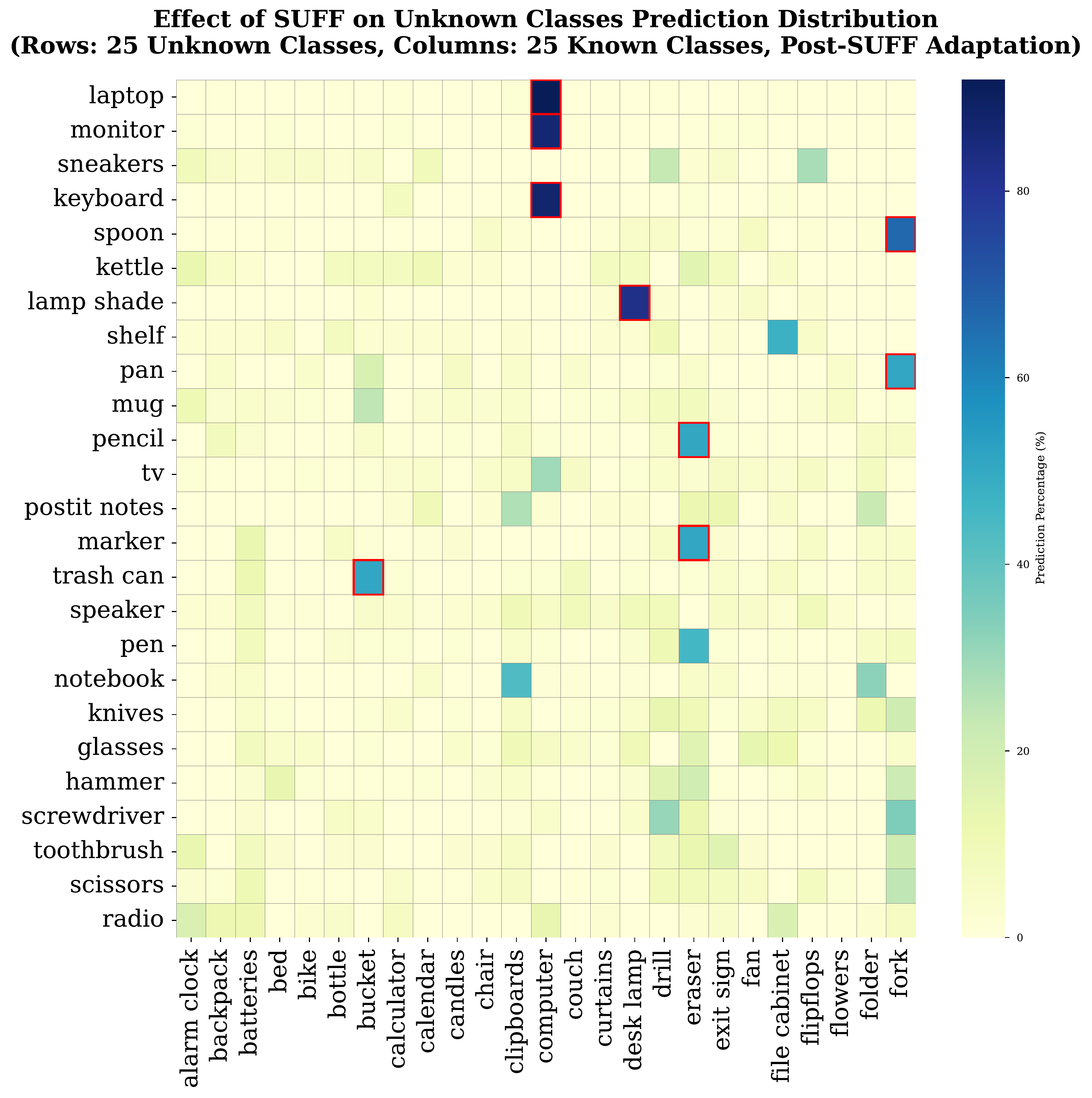}}
    \hfill
    \subfloat[Product (CLIP + SUFF)]{\includegraphics[width=0.23\textwidth]{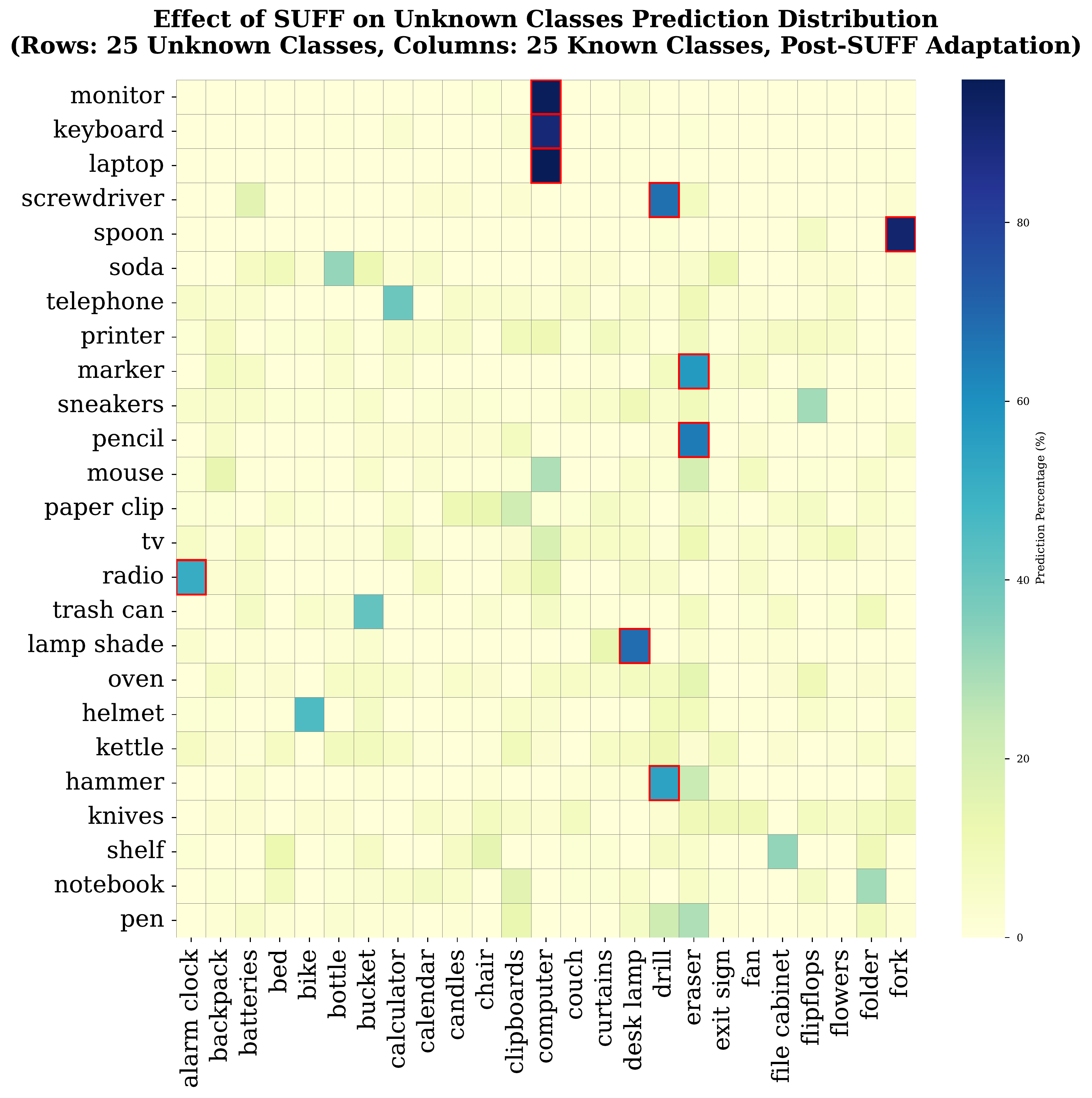}}
    \hfill
    \subfloat[Real World (CLIP + SUFF)]{\includegraphics[width=0.23\textwidth]{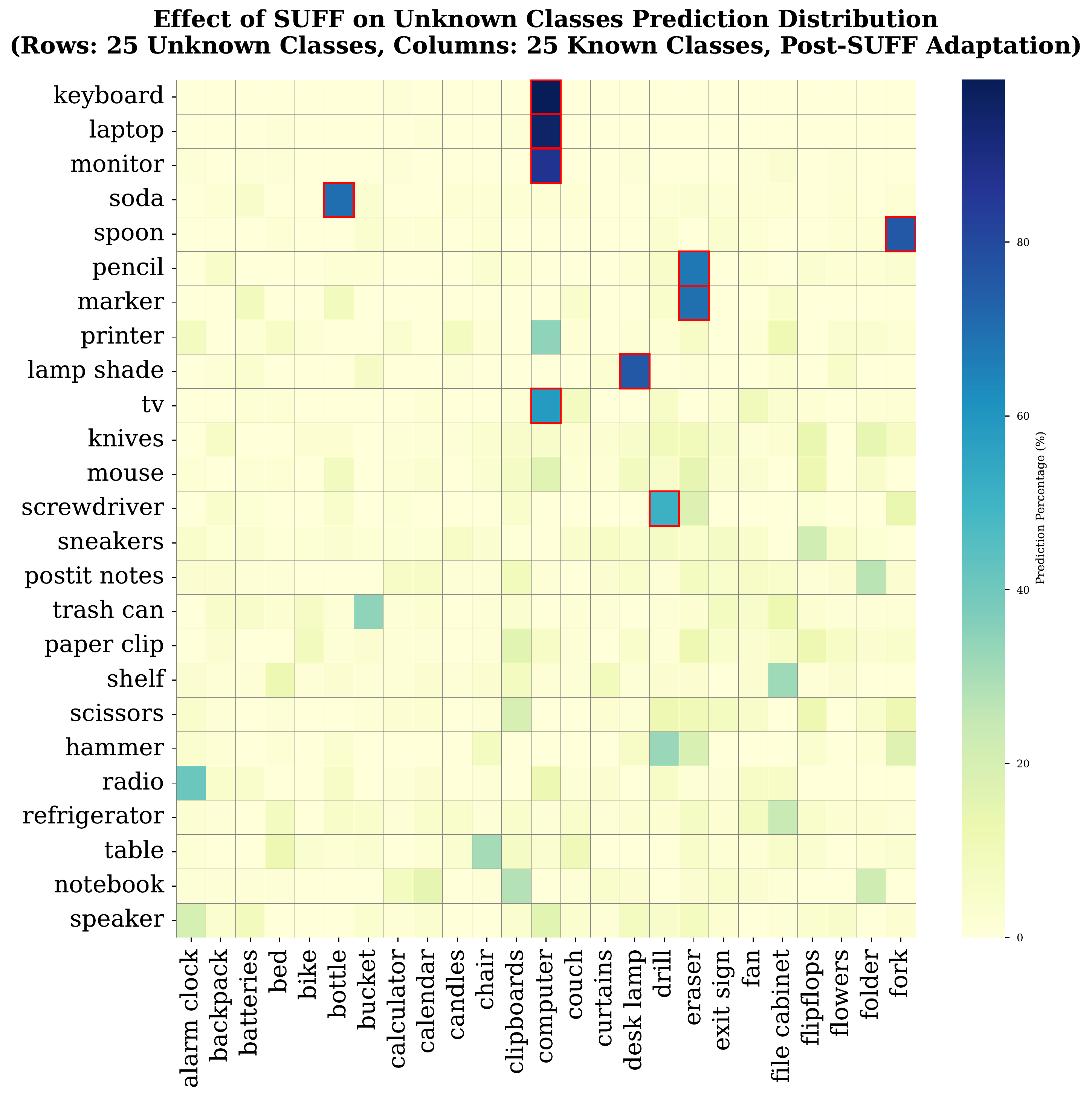}}

    \caption{Samples of the 25 classes with the strongest tendency towards known classes from the four domains of the Office-Home dataset. The figure shows the changes in classification results for known classes between the original CLIP features and the features filtered by the SUFF module. Rows represent unknown classes, columns represent known classes, and the darker the cell, the more samples from the corresponding unknown class are classified into the corresponding known class.}
    \label{fig:appendix_detail_confusion}
\end{figure*}

\begin{figure*}[ht]
    \centering
    \subfloat[Art (CLIP)]{\includegraphics[width=0.23\textwidth]{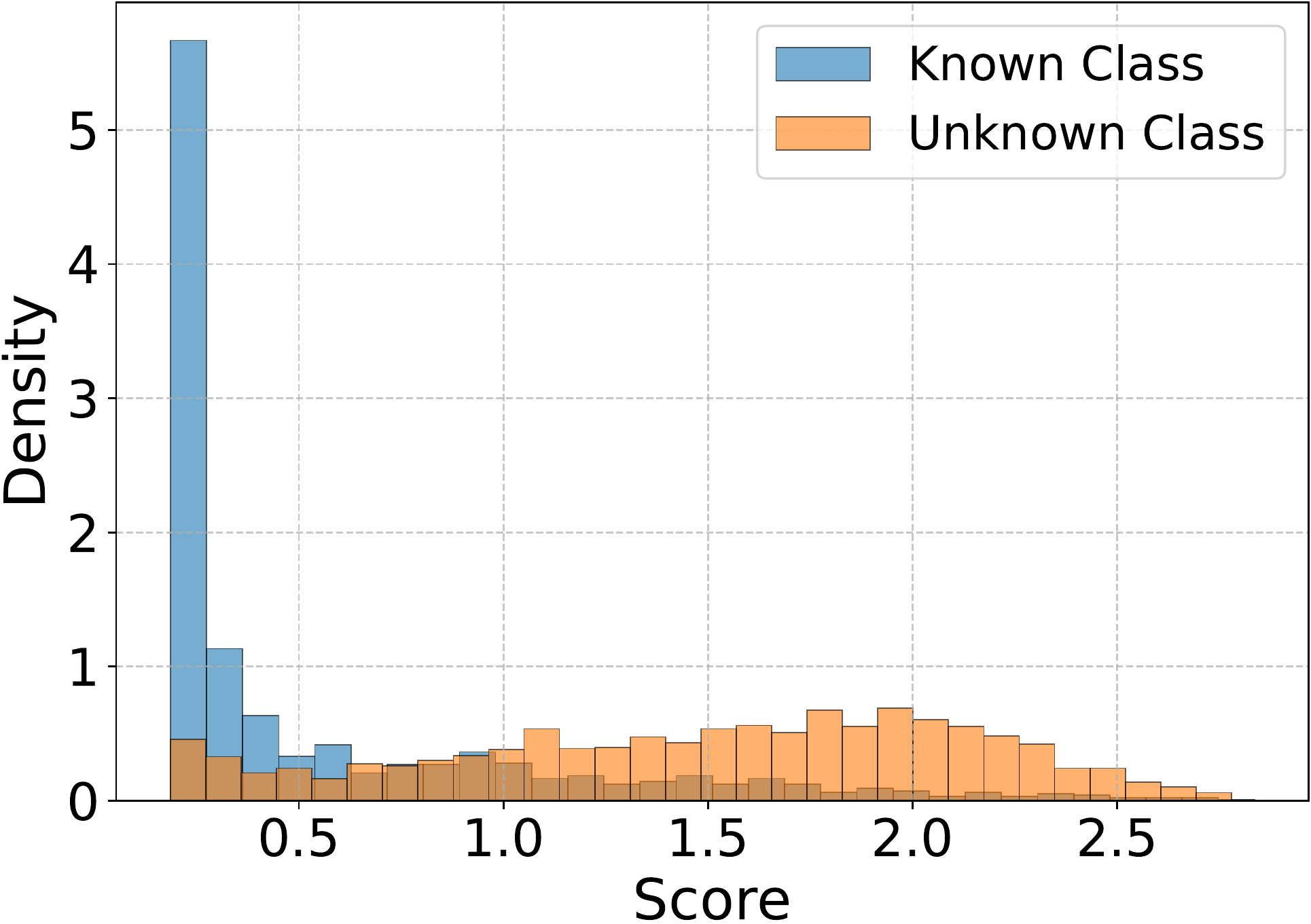}}
    \hfill
    \subfloat[Clipart (CLIP)]{\includegraphics[width=0.23\textwidth]{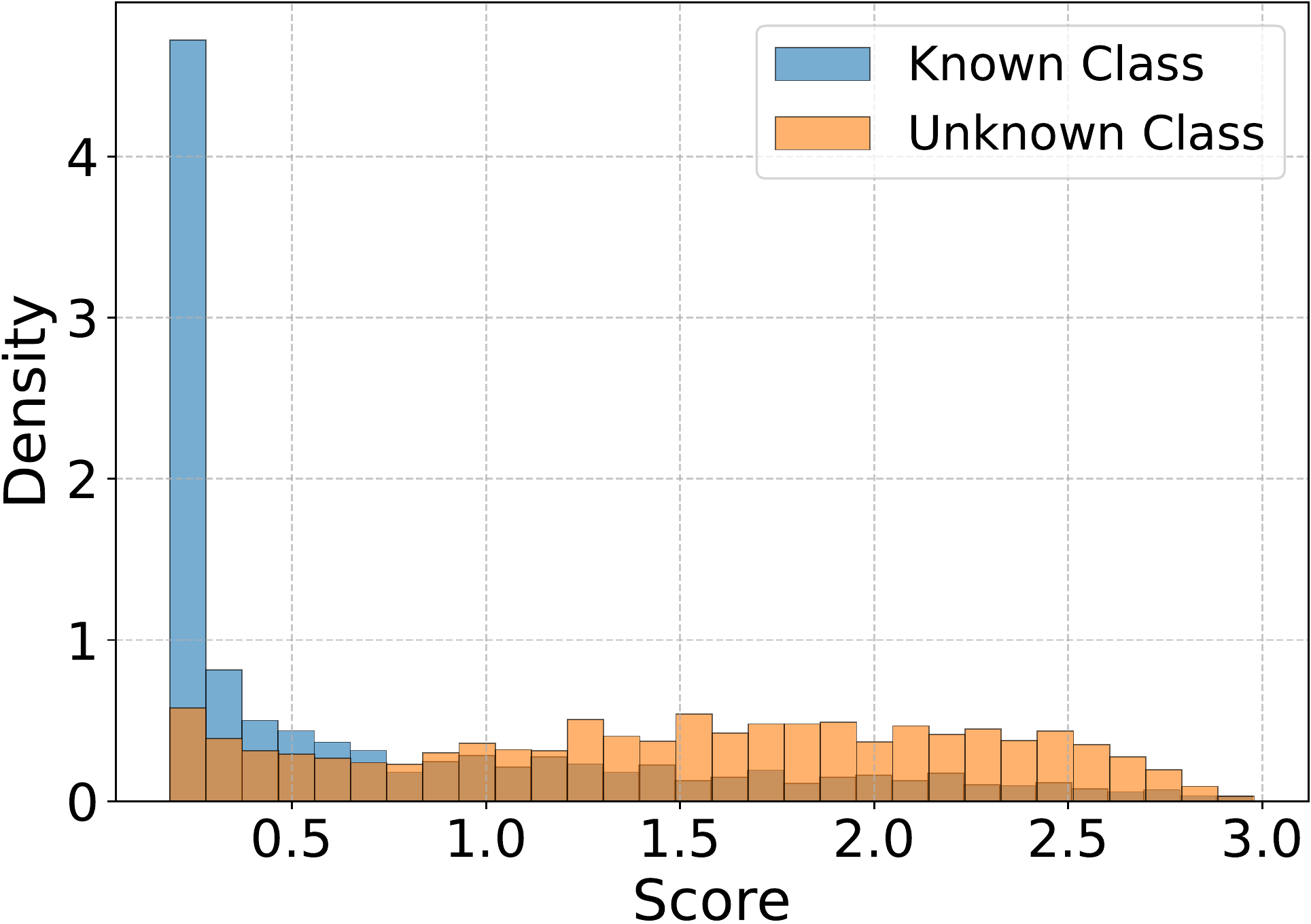}}
    \hfill
    \subfloat[Product (CLIP)]{\includegraphics[width=0.23\textwidth]{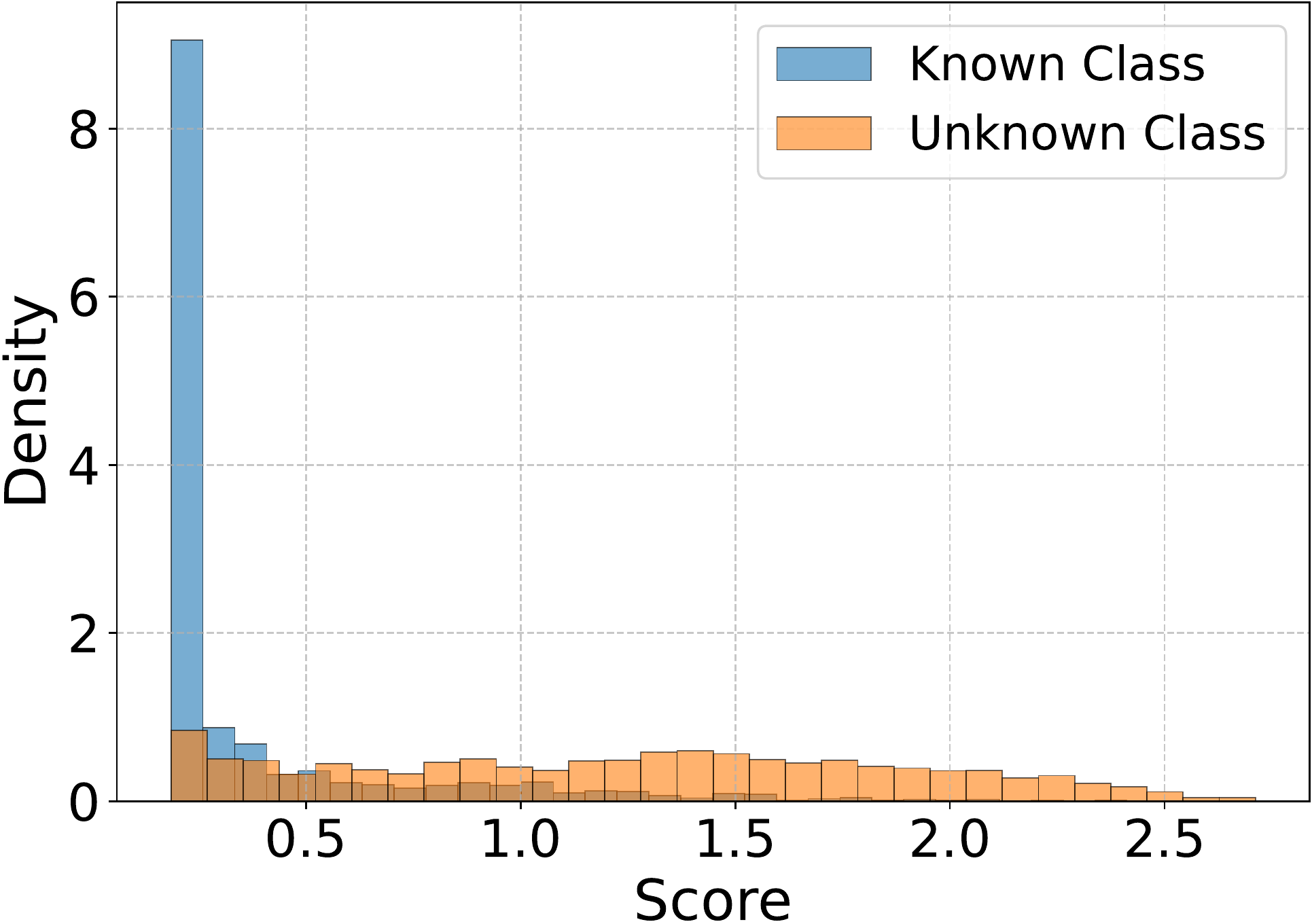}}
    \hfill
    \subfloat[Real World (CLIP)]{\includegraphics[width=0.23\textwidth]{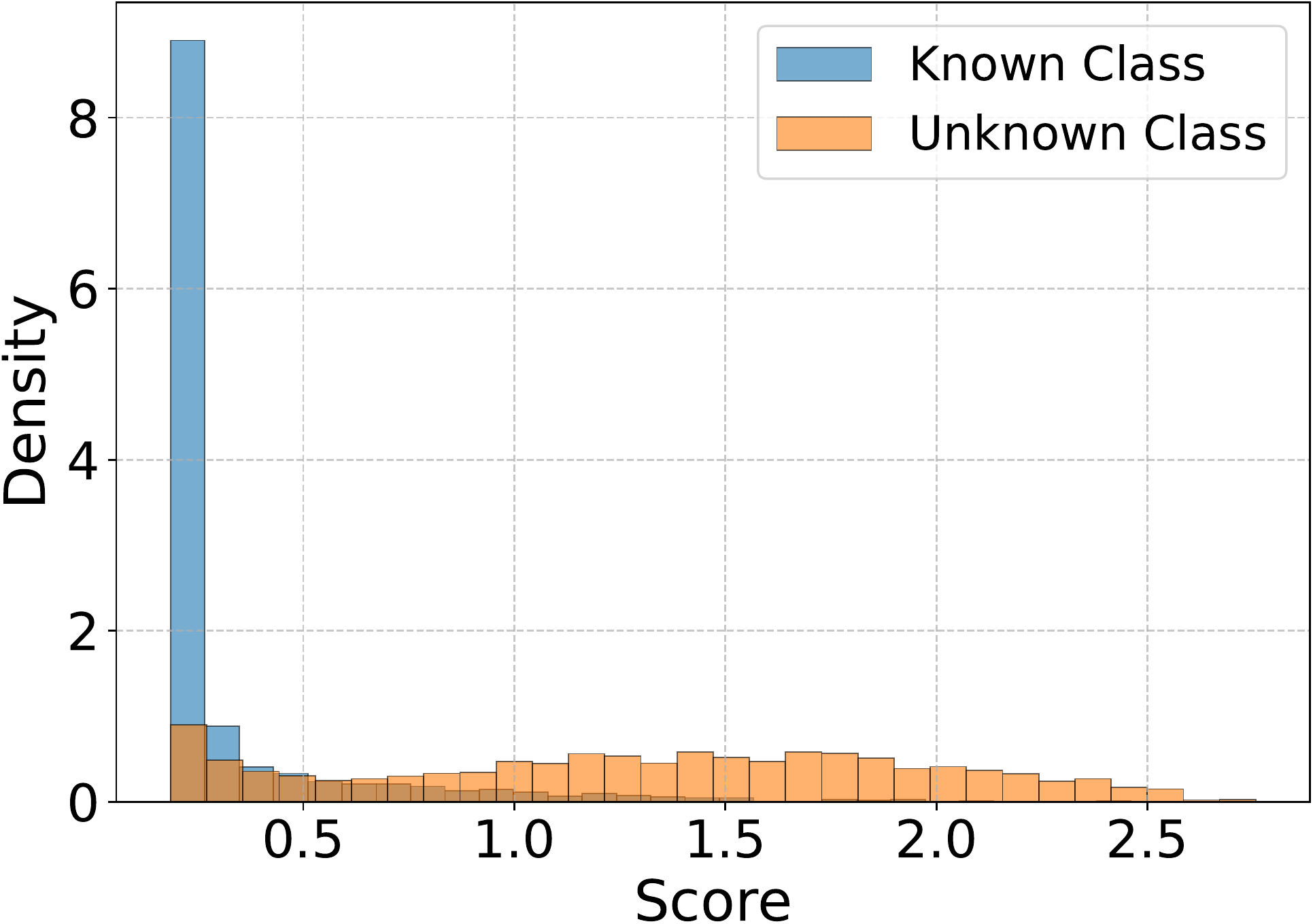}}
    
    \vspace{5pt}
    
    \subfloat[Art (CLIP + SUFF)]{\includegraphics[width=0.23\textwidth]{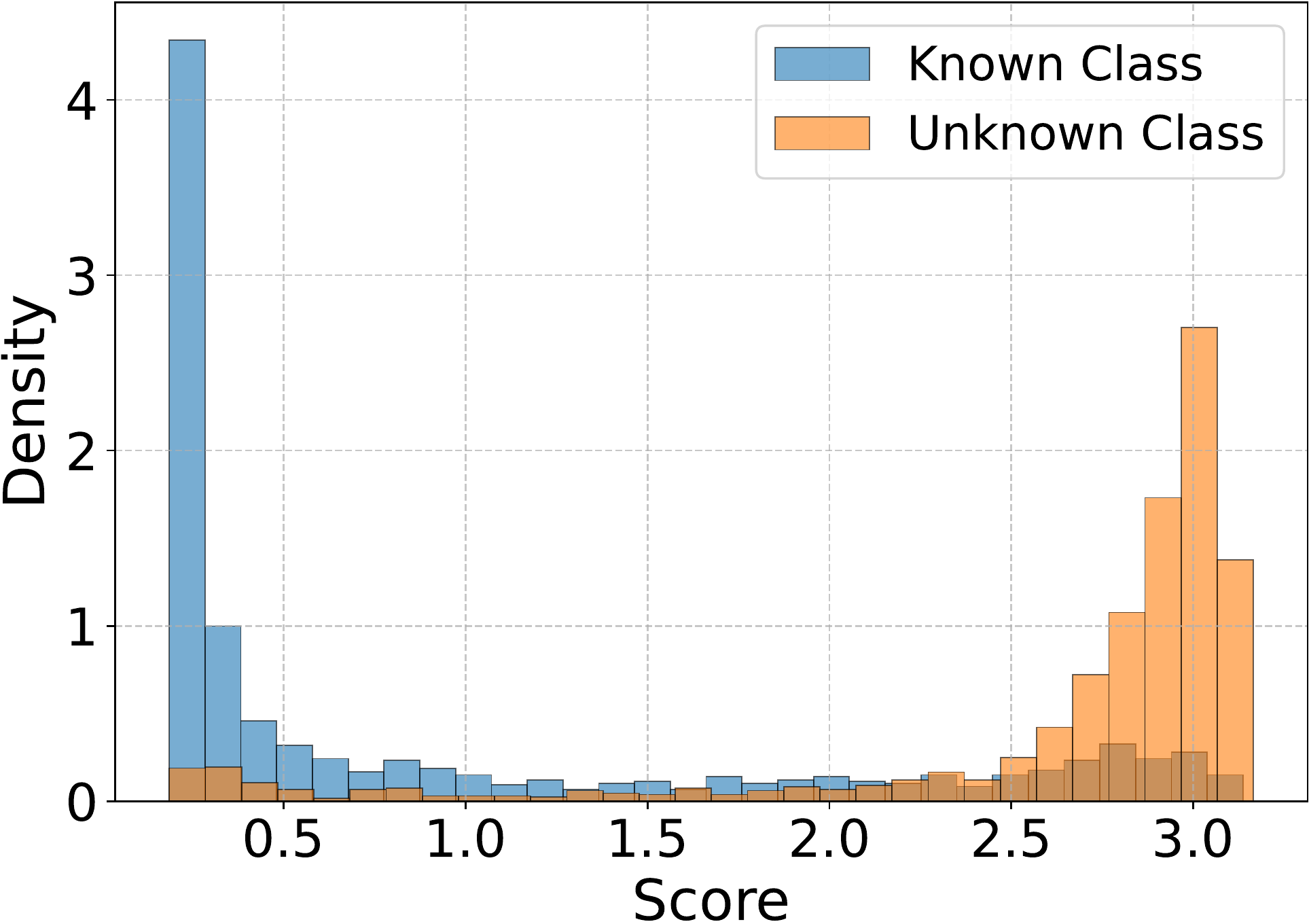}}
    \hfill
    \subfloat[Clipart (CLIP + SUFF)]{\includegraphics[width=0.23\textwidth]{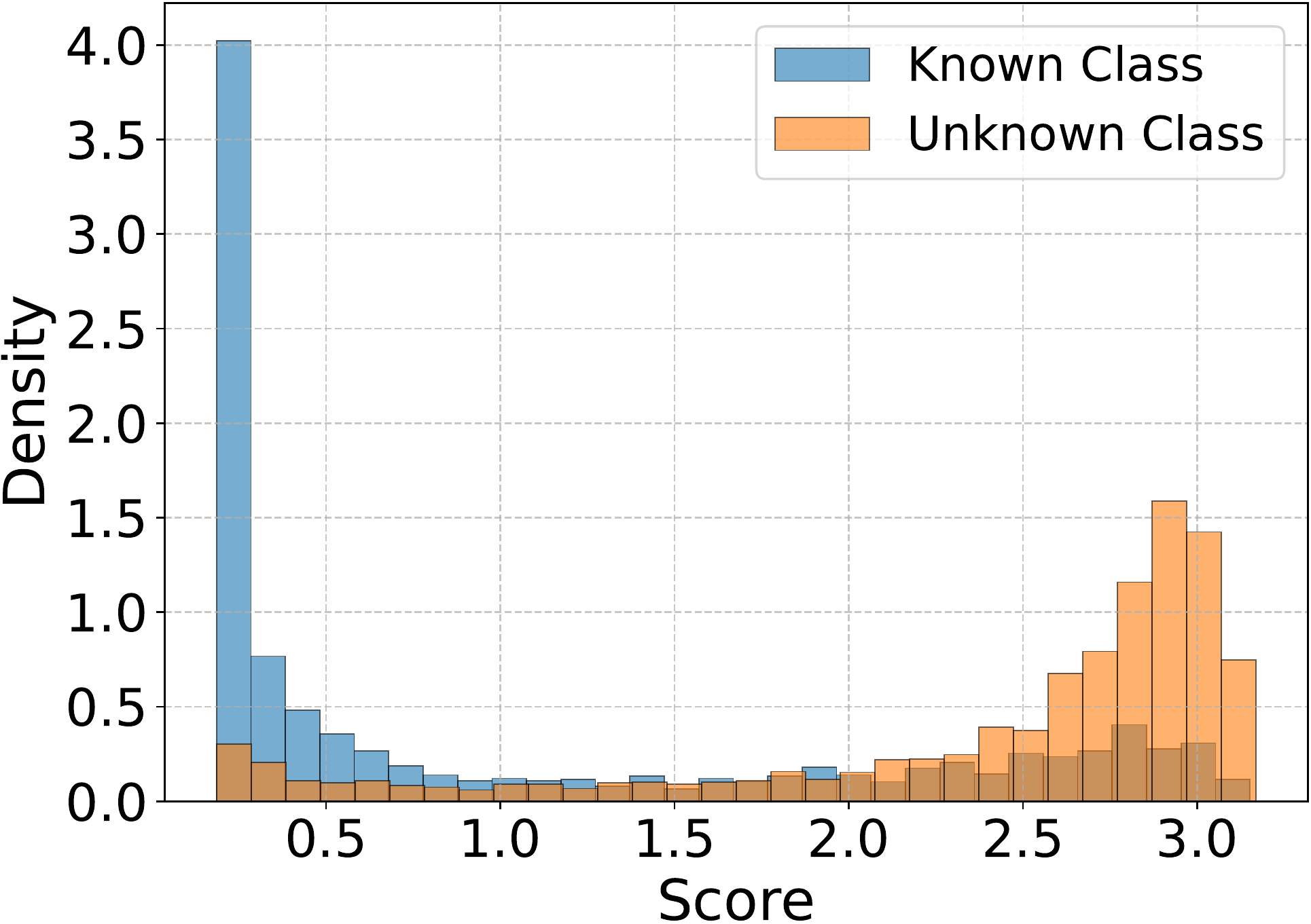}}
    \hfill
    \subfloat[Product (CLIP + SUFF)]{\includegraphics[width=0.23\textwidth]{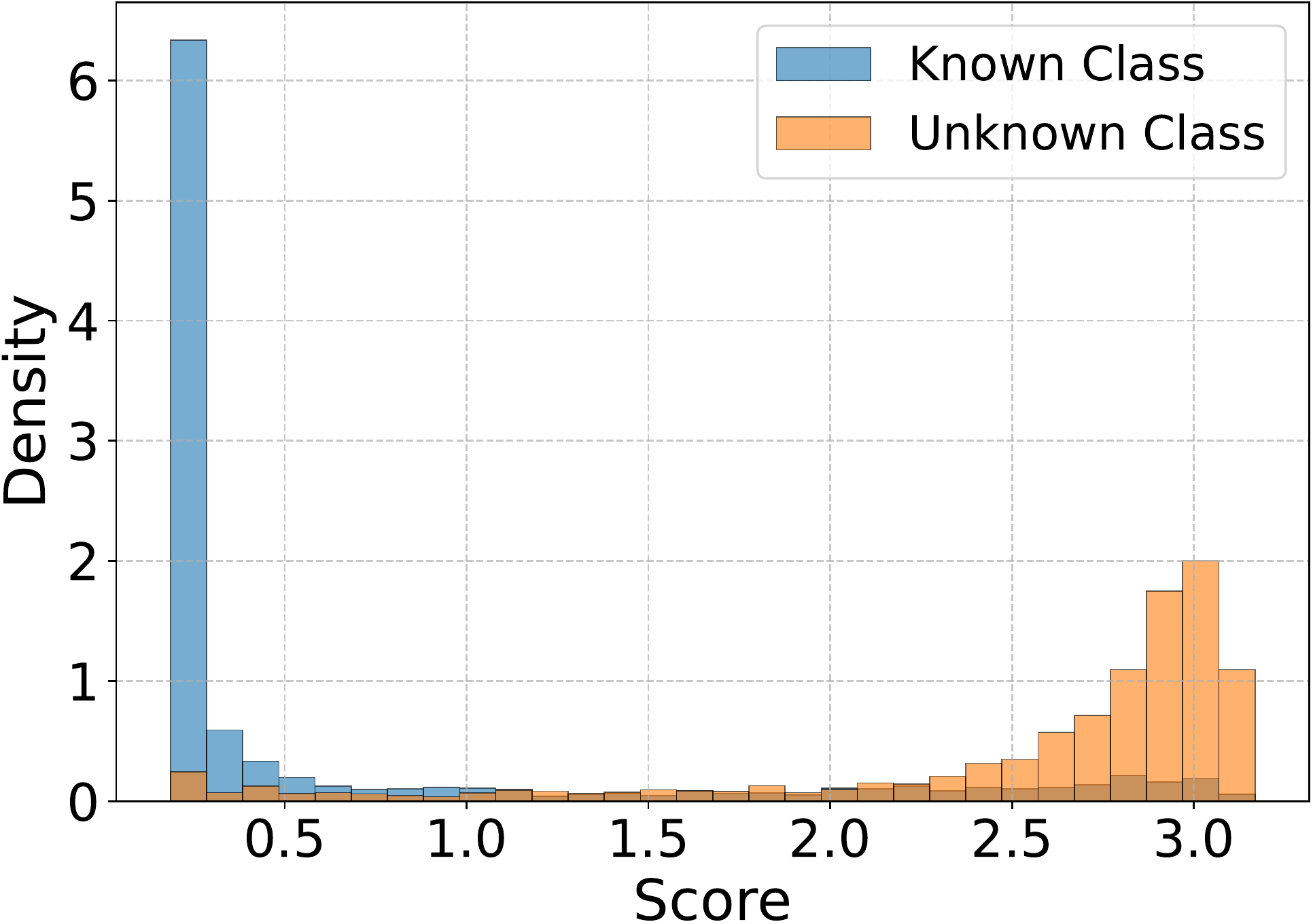}}
    \hfill
    \subfloat[Real World (CLIP + SUFF)]{\includegraphics[width=0.23\textwidth]{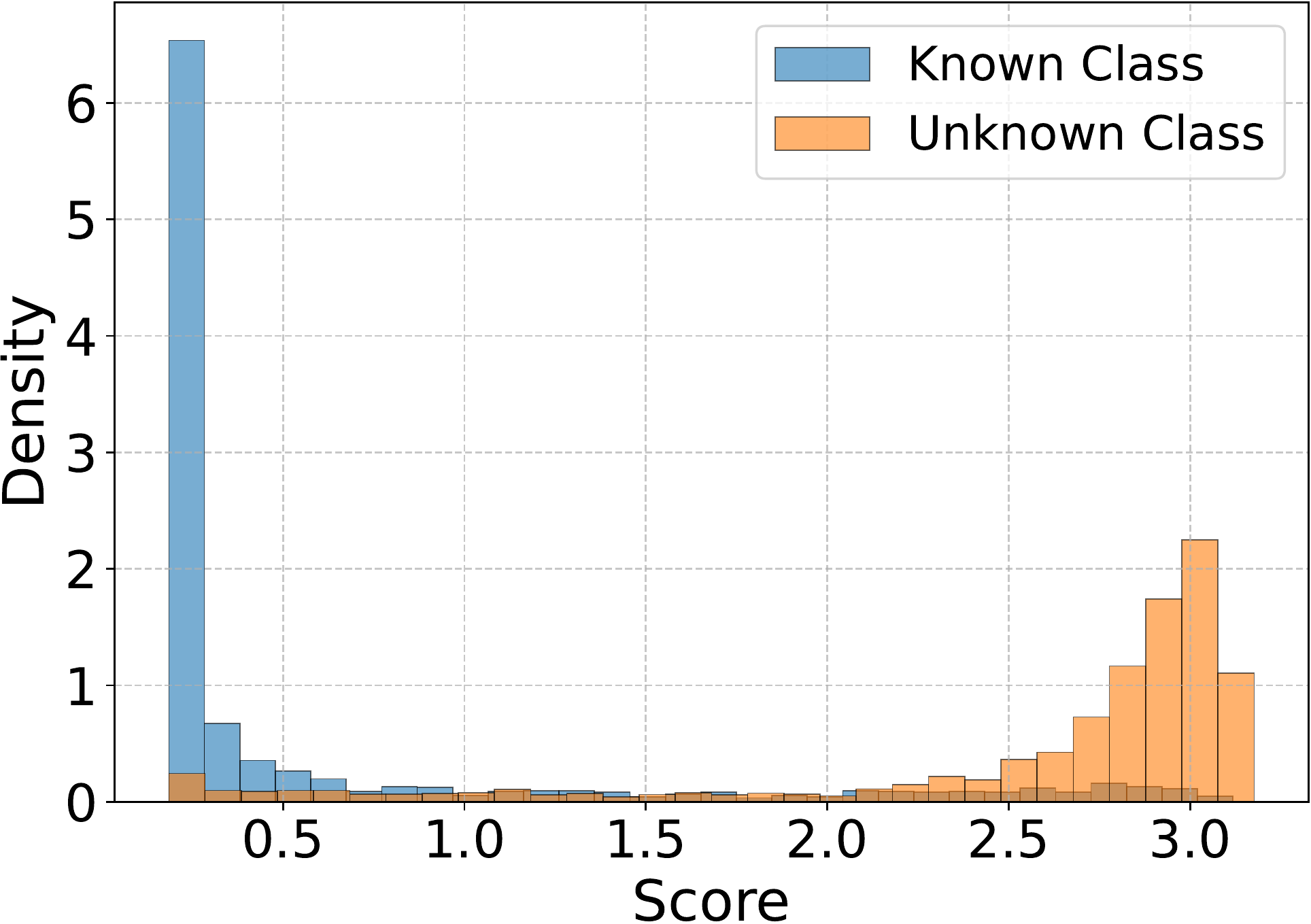}}

    \caption{Changes in the entropy of samples from the four domains of the Office-Home dataset before and after processing by the SUFF module. The first row shows the entropy computed from the original CLIP features, while the second row displays the entropy of the samples after filtering by the SUFF module.}
    \label{fig:appendix_detail_entropy_his}
\end{figure*}

\textbf{Analysis of the SAA Phenomenon. }Since vision–language models (VLMs) perform classification by measuring similarity between image and text embeddings, the predicted label distribution of unknown samples reflects their semantic proximity to known categories. 

In an ideal open-set setting, if the unknown classes are semantically unrelated to all known classes, their predictions should be uniformly scattered across the known categories due to the absence of strong similarity signals. However, when certain unknown classes are semantically close to specific known classes, VLMs tend to consistently misclassify them into those nearby categories, forming a phenomenon we refer to as \textit{semantic affinity anchoring}.

To quantitatively assess the prevalence of this Semantic Affinity Anchoring (SAA) effect in existing benchmarks, we propose a simple yet effective metric: \textbf{Anchor Rate at Top-1 (AR@1)}. 
This metric measures, at the dataset level, the proportion of unknown classes whose samples are consistently mapped to a particular known class.

Formally, we define the following:
\begin{itemize}
  \item $C$ be the known‑class set, $|C|=K$;
  \item $U$ the unknown classes that actually occur in the test set, $|U|=M$;
  \item $N_u$ the number of test samples belonging to the unknown class $u\in U$;
  \item $\hat y_i\in C$ the VLM’s top‑1 prediction for sample $x_i$.
\end{itemize}

\begin{table}[htbp]
  \centering
  \renewcommand{\arraystretch}{1.2}
  \resizebox{\linewidth}{!}{
    \begin{tabular}{c|c|c|cccc|c|c}
    \toprule[1.5pt]
    \multirow{2}[0]{*}{Method} & \multirow{2}[0]{*}{Treshold} & \multirow{2}[0]{*}{ImageNet} & \multicolumn{4}{c|}{OfficeHome} & \multirow{2}[0]{*}{VisDA} & \multirow{2}[0]{*}{AVG} \\ \cline{4-7}
          &       &       & Art   & Clipart & Product & \multicolumn{1}{c|}{Real}  &       &  \\ \hline
          \multirow{5}[0]{*}{CLIP}& Fixed & 61.6  & 61.9  & 59.9  & 46.4  & 52.8  & 83.8  & 61.1  \\
          & Mean  & 61.9  & 76.4  & 69.6  & 76.1  & 80.1  & 86.8  & 75.1  \\
          & K-means & 62.1  & 74.7  & 69.3  & 72.7  & 77.4  & 86.3  & 73.8  \\
          & GMM-int & 63.2  & 57.5  & 54.8  & 58.7  & 57.7  & 72.1  & 60.7  \\
          & BGAT  & \textbf{63.3}  & \textbf{76.5}  & \textbf{71.7}  & \textbf{82.1}  & \textbf{82.8}  & \textbf{87.8}  & \textbf{77.4}  \\
          \bottomrule[1.5pt]
    \end{tabular}%
    }
      \caption{Comparison of HOS (\%) for the CLIP model under different threshold functions.
}
  \label{tab:appendix_detail_clip_threshold}%
\end{table}%


For each unknown class \(u\), we build a prediction distribution over known classes as:
\[
p_{u,c}
  \;=\;
  \frac{1}{N_u}\sum_{i:\,y_i=u}
  \mathbf 1[\hat y_i = c],
\qquad
\sum_{c\in C} p_{u,c}=1 .
\]

Given a threshold \(\tau\in(0,1)\), the Anchor Rate at Top‑1 is:
\[
\mathrm{AR@1}(\tau)
\;=\;
\frac{1}{M}
\sum_{u\in U}
  \mathbf 1\!\bigl[
    \max_{c\in C} p_{u,c}\;\ge\;\tau
  \bigr].
\]

The AR@1 metric possesses the following key properties:
\begin{itemize}
  \item \textbf{Dataset-level metric}: used to evaluate the proportion of unknown classes anchored to known classes in a dataset.
  \item \textbf{Threshold‑controlled}: larger \(\tau\) makes anchoring harder to satisfy. In this paper, we set $\tau = 0.5$ by default.

  \item \textbf{Range}: \(\mathrm{AR@1}\in[0,1]\); A value closer to 0 indicates fewer unknown classes are anchored, while a value closer to 1 indicates more anchoring. A value of 1 means all unknown classes are anchored to specific known classes, whereas 0 means none exhibit significant anchoring.
\end{itemize}

As reported in Table~\ref{tab:detailed_AR1}, all three VLMs demonstrate a pronounced semantic affinity anchoring (SAA) effect across all datasets.
 It can be observed that, across all three VLM models, the average AR@1 scores exceed 0.5, indicating that over 50\% of the unknown classes have more than half of their samples anchored to the same known class. This demonstrates that the Semantic Affinity Anchoring (SAA) phenomenon is prevalent in current benchmark datasets.


\begin{table}[htbp]
  \centering
  \renewcommand{\arraystretch}{1.2}
  \resizebox{\linewidth}{!}{
    \begin{tabular}{c|c|c|cccc|c|c}
    \toprule[1.5pt]
    \multirow{2}[0]{*}{Method} & \multirow{2}[0]{*}{Treshold} & \multirow{2}[0]{*}{ImageNet} & \multicolumn{4}{c|}{OfficeHome} & \multirow{2}[0]{*}{VisDA} & \multirow{2}[0]{*}{AVG} \\ \cline{4-7}
          &       &       & Art   & Clipart & Product & \multicolumn{1}{c|}{Real}  &       &  \\ \hline
          \multirow{5}[0]{*}{CLIP w/ SUFF}& Fixed & 69.6  & 80.2  & 72.1  & 85.7  & 85.0  & 89.0  & 80.3  \\
          & Mean  & 70.2  & \textbf{80.4}  & 72.3  & 84.2  & 84.7  & 88.3  & 80.0  \\
          & K-means & 70.8  & 79.9  & 73.1  & 86.2  & 85.2  & 89.0  & 80.7  \\
          & GMM-int & 69.3  & 76.6  & 58.8  & 53.7  & 57.4  & 88.6  & 67.4  \\
          & BGAT  & \textbf{71.3}  & 78.2  & \textbf{74.0}  & \textbf{86.8}  & \textbf{85.7}  & \textbf{89.0}  & \textbf{80.8}  \\
          \bottomrule[1.5pt]
    \end{tabular}%
    }
      \caption{Comparison of HOS (\%) for the CLIP model with SUFF under different threshold functions.
}
  \label{tab:addlabel}%
\end{table}%

\textbf{Effectiveness of SUFF.}
The SUFF module is designed to address the difficulty of detecting certain known classes caused by semantic affinity anchoring (SAA). To fully assess its impact, we evaluate its performance in two ways. First, as shown in Table~\ref{tab:detailed_AR1}, adding SUFF to CLIP, SigLIP, and ALIGN reduces AR\@1 by 0.28, 30.0, and 0.31, respectively, demonstrating its effectiveness at alleviating SAA. Second, examining overall performance (Tables~\ref{tab:appendix_detail_suff_fixed}, \ref{tab:appendix_detail_suff_kmeans}, \ref{tab:appendix_detail_suff_bgat} ), SUFF significantly boosts HOS across all threshold functions—most notably by 19.2\%, 14.0\%, and 6.3\% under the simple Fixed threshold—and also delivers substantial gains under the K-meas and our BGAT thresholds. These results confirm that SUFF markedly improves model performance by mitigating the SAA effect.

\begin{table}[htbp]
  \centering
    \renewcommand{\arraystretch}{1.2}
  \resizebox{\linewidth}{!}{
    \begin{tabular}{c|c|c|cccc|c|c}
    \toprule[1.5pt]
    \multirow{2}[0]{*}{Method} & \multirow{2}[0]{*}{Treshold} & \multirow{2}[0]{*}{ImageNet} & \multicolumn{4}{c|}{OfficeHome} & \multirow{2}[0]{*}{VisDA} & \multirow{2}[0]{*}{AVG} \\ \cline{4-7}
          &       &       & Art   & Clipart & Product & \multicolumn{1}{c|}{Real}  &       &  \\ \hline
         \multirow{5}[0]{*}{SigLIP} & Fixed & 52.6  & 53.0  & 42.8  & 48.7  & 47.4  & 55.6  & 50.0  \\
          & Mean  & 63.2  & 80.5  & 72.5  & 76.5  & 78.9  & 81.5  & 75.5  \\
          & K-means & 60.9  & 76.5  & 65.0  & 70.1  & 73.4  & 75.8  & 70.3  \\
          & GMM-int & 37.3  & 68.5  & 69.6  & 77.2  & 80.7  & 74.7  & 68.0  \\
          & BGAT  & \textbf{67.2}  & \textbf{83.7}  & \textbf{80.8}  & \textbf{89.2}  & \textbf{88.2}  & \textbf{87.4}  & \textbf{82.7}  \\
          \bottomrule[1.5pt]
    \end{tabular}%
    }
      \caption{Comparison of HOS (\%) for the SigLIP model under different threshold functions.}
  \label{tab:appendix_detail_siglip_threshold}%
\end{table}%

\textbf{Effectiveness of BGAT.}
We compare our BGAT adaptive‑threshold module against four established methods: a fixed threshold at half the maximum score (Fixed)~\cite{2025_CVIU_ODAwVL}, the mean of all sample scores (Mean)~\cite{2022_ECCV_ROS_OSDA}, k‑means clustering (K‑Means)~\cite{2020_ICML_SHOT}, and the intersection point of two Gaussian densities (GMM‑Int)~\cite{2025_AAAI_TASC}.  We conduct evaluation across six diverse tasks—ImageNet, four domains from OfficeHome (Art, Clipart, Product, Real), and VisDA-2017—and test three different VLM models to comprehensively validate the effectiveness of our thresholding function.

\begin{table}[htbp]
  \centering
  \renewcommand{\arraystretch}{1.2}
  \resizebox{\linewidth}{!}{
    \begin{tabular}{c|c|c|cccc|c|c}
    \toprule[1.5pt]
    \multirow{2}[0]{*}{Method} & \multirow{2}[0]{*}{Treshold} & \multirow{2}[0]{*}{ImageNet} & \multicolumn{4}{c|}{OfficeHome} & \multirow{2}[0]{*}{VisDA} & \multirow{2}[0]{*}{AVG} \\ \cline{4-7}
          &       &       & Art   & Clipart & Product &  \multicolumn{1}{c|}{Real}  &       &  \\ \hline
         \multirow{5}[0]{*}{SigLIP w/ SUFF} & Fixed & 64.1  & 67.7  & 47.8  & 68.8  & 66.4  & 69.5  & 64.0  \\
          & Mean  & 66.7  & 79.2  & 70.9  & 77.7  & 78.8  & 81.5  & 75.8  \\
          & K-means & 66.5  & 80.6  & 67.8  & 80.8  & 81.3  & 80.3  & 76.2  \\
          & GMM-int & 44.8  & 79.4  & \textbf{80.9}  & 84.7  & 84.1  & 80.8  & 75.8  \\
          & BGAT  & \textbf{69.9}  & \textbf{85.8}  & 80.4  & \textbf{89.4}  & \textbf{88.9}  & \textbf{88.1}  & \textbf{83.7}  \\
          \bottomrule[1.5pt]
    \end{tabular}%
    }
      \caption{Comparison of HOS (\%) for the SigLIP model with SUFF under different threshold functions.}
  \label{tab:appendix_detail_siglip_suff_threshold}%
\end{table}%

As shown in Tables~\ref{tab:appendix_detail_clip_threshold}-\ref{tab:appendix_detail_align_SUFF_threshold}, our BGAT module consistently delivers the best performance across all VLM backbones. In particular, on the raw scores derived from pre‑SUFF features , BGAT outperforms the four baseline thresholding methods on SigLIP by 32.7\%, 7.2\%, 12.4\%, and 14.7\%, respectively. This gap arises because the original VLM score distributions exhibit poor class separability and are highly sensitive to threshold choice. Even after applying the SUFF module—which substantially reduces score sensitivity—BGAT still surpasses the competing methods by 19.7\%, 7.9\%, 7.5\%, and 7.9\%. These results confirm that BGAT provides a more robust and reliable threshold selection across a wide range of score distributions.

\textbf{Effectiveness of Box-Cox.}
The BGAT module consists of a Box-Cox transformation and a Gaussian mixture model. The Box-Cox transformation aims to correct the skewness of the score distribution, making it more Gaussian-like and improving the accuracy of threshold estimation. To validate its effectiveness, we conduct experiments on all four domains of Office-Home, comparing the performance of CLIP, SigLIP, and ALIGN with and without the Box-Cox transformation.

\begin{table}[htbp]
  \centering
    \renewcommand{\arraystretch}{1.2}
  \resizebox{\linewidth}{!}{
    \begin{tabular}{c|c|c|cccc|c|c}
    \toprule[1.5pt]
    \multirow{2}[0]{*}{Method} & \multirow{2}[0]{*}{Treshold} & \multirow{2}[0]{*}{ImageNet} & \multicolumn{4}{c|}{OfficeHome} & \multirow{2}[0]{*}{VisDA} & \multirow{2}[0]{*}{AVG} \\ \cline{4-7}
          &       &       & Art   & Clipart & Product & \multicolumn{1}{c|}{Real}  &       &  \\ \hline
         \multirow{5}[0]{*}{ALIGN} & Fixed & 57.1  & 78.4  & 73.1  & 89.9  & 79.9  & 75.1  & 75.6  \\
          & Mean  & 59.3  & 82.6  & 77.9  & 90.0  & 86.1  & 83.0  & 79.8  \\
          & K-means & 52.6  & \textbf{81.7}  & 77.0  & 90.6  & 84.7  & 75.8  & 77.1  \\
          & GMM-int & 59.0  & 57.2  & 66.4  & 52.9  & 67.3  & 66.2  & 61.5  \\
          & BGAT  & \textbf{59.4}  & 81.4  & \textbf{81.1}  & \textbf{90.8}  & \textbf{88.9}  & \textbf{86.1}  & \textbf{81.3}  \\ \bottomrule[1.5pt]
    \end{tabular}%
    }
    \caption{Comparison of HOS (\%) for the ALIGN model under different threshold functions.}
  \label{tab:appendix_detail_align_threshold}%
\end{table}%

\begin{table}[htbp]
    \centering
    \renewcommand{\arraystretch}{1.2}
  \resizebox{\linewidth}{!}{
    \begin{tabular}{c|c|c|cccc|c|c}
    \toprule[1.5pt]
    \multirow{2}[0]{*}{Method} & \multirow{2}[0]{*}{Treshold} & \multirow{2}[0]{*}{ImageNet} & \multicolumn{4}{c|}{OfficeHome} & \multirow{2}[0]{*}{VisDA} & \multirow{2}[0]{*}{AVG} \\ \cline{4-7}
          &       &       & Art   & Clipart & Product & \multicolumn{1}{c|}{Real}  &       &  \\ \hline
         \multirow{5}[0]{*}{ALIGN w/ SUFF} & Fixed & 59.6  & 84.4  & 80.7  & 90.5  & 90.3  & 86.1  & 81.9  \\
          & Mean  & 61.4  & 84.3  & 81.3  & 91.0  & 87.7  & 86.2  & 82.0  \\
          & K-means & \textbf{64.1}  & 84.5  & 82.2  & 90.9  & 90.9  & 80.3  & 82.1  \\
          & GMM-int & 53.1  & 59.8  & 67.4  & 86.6  & 63.7  & 66.3  & 66.1  \\
          & BGAT  & 60.7  & \textbf{85.0}  & \textbf{83.2}  & \textbf{90.9}  & \textbf{91.3}  & \textbf{87.3}  & \textbf{83.1}  \\
      \bottomrule[1.5pt]
    \end{tabular}%
    }
    \caption{Comparison of HOS (\%) for the ALIGN model with SUFF under different threshold functions.}
  \label{tab:appendix_detail_align_SUFF_threshold}%
\end{table}%

\begin{table}[htbp]
  \centering
      \renewcommand{\arraystretch}{1.0}
    \resizebox{\linewidth}{!}{
    \begin{tabular}{c|c|cccc|c|c}
        \toprule[1.5pt]
    \multirow{2}[0]{*}{VLM} & \multirow{2}[0]{*}{Box-Cox} & \multicolumn{4}{c|}{Office-Home} & \multirow{2}[0]{*}{VisDA} & \multirow{2}[0]{*}{AVG} \\ \cline{3-6}
          &       & Art   & Clipart & Product & Real  &       &  \\  \hline
    \multirow{2}[0]{*}{SigLIP} & no    & 81.5  & 72.9  & 80.7  & 80.9  & 83.6  & 79.9  \\
          & yes   & \textbf{85.8}  & \textbf{80.4}  & \textbf{89.4}  & \textbf{88.9}  & \textbf{88.1}  & \textbf{86.5}  \\ \hline
    \multirow{2}[0]{*}{ALIGN} & no    & 84.4  & 82.0  & \textbf{91.3}  & 89.1  & 86.9  & 86.7  \\
          & yes   & \textbf{85.0}  & \textbf{83.2}  & 90.9  & \textbf{91.3}  & \textbf{87.3}  & \textbf{87.5}  \\  \hline
    \multirow{2}[0]{*}{CLIP} & no    & 77.6  & 73.9  & 85.8  & 84.7  & 88.6  & 82.1  \\
          & yes   & \textbf{78.2}  & \textbf{74.0}  & \textbf{86.8}  & \textbf{85.7}  & \textbf{89.0}  & \textbf{82.7}  \\
          \bottomrule[1.5pt]
    \end{tabular}%
    }
    \caption{ HOS (\%) before and after applying the Box–Cox transformation across different VLMs. }
  \label{tab:appendix_box_cox}%
\end{table}%

\begin{figure}[]
    \centering
    \begin{minipage}{0.47\linewidth}
        \centering
        \includegraphics[width=\linewidth]{figs/CLIP_Art_score2.pdf}
        \label{fig:clip_entropy_distribution}
    \end{minipage}%
    \hfill
    \hspace{0.2em}
    \begin{minipage}{0.49\linewidth}
        \centering
        \includegraphics[width=\linewidth]{figs/SigLIP_Art_score2.pdf}
        \label{fig:siglip_entropy_distribution}
    \end{minipage} 
\caption{Entropy distribution of sample scores from CLIP and SigLIP on the Art domain of the Office-Home dataset.
}
\label{fig:VLM_entropy_distribution}
\end{figure}

As shown in Table~\ref{tab:appendix_box_cox}, the Box-Cox transformation is highly effective for the SigLIP model, yielding an average performance gain of 6.6\% across the four domains. In contrast, CLIP and ALIGN achieve smaller improvements of 0.6\% and 0.8\%, respectively, with an overall average gain of 2.67\%. This demonstrates the effectiveness of the proposed Box-Cox transformation. Moreover, the larger gain observed in SigLIP can be attributed to its score distribution: as shown in Figure~\ref{fig:VLM_entropy_distribution}, the known-class scores (measured by entropy) in SigLIP exhibit significantly higher skewness than those in CLIP on the same dataset.

\subsection{Runtime Analysis and Comparison}
\label{appendix:time_analyse}
\textbf{Runtime Profiling of Inference and SUFF (SVD) Module.}
Our VLM‑OpenXpert framework adds two inference modules that introduce a modest overhead to the base model’s runtime. To quantify this overhead, we profiled the time spent in each component on the training splits of the Real domain in Office‑Home, the VisDA‑2017 benchmark, and all domains of DomainNet. All measurements were taken under identical hardware conditions: an Intel Xeon Gold 6442Y CPU and a Tesla L40 GPU.

We partitioned the inference process into three components: (1) VLM feature encoding; (2) the two SVD decompositions in SUFF; and (3) all other feature operations and processing, including BGAT. As shown in Table~\ref{tab:time_analyse}, roughly 97.5\% of runtime is spent on VLM feature encoding, SVD takes under 0.5\%, and the remaining operations (including BGAT) account for only about 2\%. This confirms that our method adds minimal computational overhead to the inference process.

\textbf{Adaptation Time Comparison with Trainable SF-OSDA Methods.}
We conduct this experiment to evaluate the adaptation time advantage of our method over trainable SF-OSDA approaches. Specifically, we compare with two representative methods: SHOT, a widely used SF-OSDA method, and UEO, a target-only open-set adaptation method based on the same VLM (CLIP). For a fair comparison, we only measure the adaptation time on the target domain, excluding source domain training. We also ensure that the image encoder is the same across methods, and that our method and UEO use the same VLM.

As shown in Table \ref{tab:time_comparision}, our method achieves significantly faster adaptation compared to other approaches. Specifically, it requires only 1.8\% of SHOT's adaptation time on the target domain. Even compared to UEO, which benefits from prompt learning and has relatively fast training, our method takes only 4.2\% of its time. These results clearly demonstrate the substantial speed advantage of our approach over previous training-intensive methods.

\textbf{Impact of Varying Unknown‐Class Proportions.}
To assess the effect of varying unknown‐class ratios, we used the Office-Home dataset across its four domains. Since this factor is seldom studied, we reproduced two representative open-source baselines—SHOT and UEO—and adopted the same CLIP backbone as UEO.

Office-Home comprises 65 classes. We evaluated five scenarios with 60, 50, 40, 30, and 25 unknown classes. Thanks to our plug-and-play design, we also measured how integrating our method into SHOT and UEO affects their performance. (For UEO as a baseline, thresholds were set to the mean score.)

As shown in Figure~\ref{fig:ueo_shot}, our method outperforms SHOT and UEO at all ratios, with peak gains of 25.5\% and 20.7\%, respectively. When our method is applied to extend SHOT and UEO, their performance further improves by up to 12.95\% under varying ratios. These results confirm the robustness of our approach to changes in the number of unknown classes.

\begin{figure}[t]
  \vspace{-1em}
  \centering
  \begin{subfigure}[b]{0.32\linewidth}
    \centering
    \includegraphics[width=\linewidth]{tabs/APPENDIX/class_ablation.pdf}
    \caption{\tiny our vs. UEO/SHOT}
    \label{fig:radar1}
  \end{subfigure}
  \hfill
  \begin{subfigure}[b]{0.32\linewidth}
    \centering 
    \includegraphics[width=\linewidth]{tabs/APPENDIX/ueo_radar.pdf}
    \caption{\tiny UEO vs. UEO with our }
    \label{fig:radar2}
  \end{subfigure}
  \hfill
  \begin{subfigure}[b]{0.32\linewidth}
    \centering
    \includegraphics[width=\linewidth]{tabs/APPENDIX/shot_radar.pdf}
    \caption{\tiny Shot vs. Shot with our}
    \label{fig:radar3}
  \end{subfigure}
  \caption{Performance vs. Number of Unknown Classes.}
  \label{fig:ueo_shot}
  \vspace{-0.5em}
\end{figure}

\begin{table}[htbp]
  \centering
   \renewcommand{\arraystretch}{1.2}
  \resizebox{\linewidth}{!}{
    \begin{tabular}{c|c|c|c}
    \toprule[1.5pt]
          & Office-Home(R) & VisDA-2017 & DomainNet \\ \hline
    Sample couts & 4375  & 55388 & 596006 \\
    Total time & 20.18 & 140.75 & 1455.82 \\
    CLIP encoding time & 19.82 & 136.29 & 1438.47 \\
    SVD time & 0.11 (0.5\%)  & 0.18 (0.1\%)  & 2.357 (0.2\%) \\
    other & 0.36 (1.8\%)  & 4.46 (3.2\%)  & 17.35 (1.2\%) \\
    \bottomrule[1.5pt]
    \end{tabular}%
    }
      \caption{Detailed time consumption of each inference component}
  \label{tab:time_analyse}%
\end{table}%

\begin{table}[htbp]
  \centering
  \renewcommand{\arraystretch}{1.2}
  \resizebox{\linewidth}{!}{
    \begin{tabular}{c|cccc|c|c}
    \toprule[1.5pt]
    \multirow{2}[0]{*}{Method} & \multicolumn{4}{c|}{OfficeHome} & \multirow{2}[0]{*}{Visda-2017} & \multirow{2}[0]{*}{AVG} \\ \cline{2-5}
          & Art   & cliipart & product & real  &       &  \\ \hline
    our-CLIP & 6.9   & 11.0  & 11.1  & 20.2  & 143.0  & 38.4  \\
    SHOT  & 325.1  & 553.6  & 566.3  & 811.2  & 8684.6  & 2188.2  \\
    UEO-CLIP   & 285.9  & 356.9  & 367.0  & 347.1  & 3192.0  & 909.8  \\
    \bottomrule[1.5pt]
    \end{tabular}%
    }
      \caption{Comparison of adaptation time across different methods}
  \label{tab:time_comparision}%
\end{table}%

\subsection{Statistical Significance and Robustness}

\begin{figure}[!t]
    \centering
    \includegraphics[width=0.99\linewidth]{figs/bgat_vs_others_forest.pdf}
    \caption{Aggregate ablation effects across 11 datasets (paired by dataset). Points denote mean HOS differences (pp) with 95\% confidence intervals; the dashed line marks no gain (0). Holm–Bonferroni–adjusted p-values and effect sizes (Cohen’s $d_z$) are annotated.}
    \label{fig:statistical_significance}
\end{figure}

\begin{table}[htbp]
  \centering
  \renewcommand{\arraystretch}{1.4}
    \resizebox{\linewidth}{!}{
    \begin{tabular}{c|c|c|c|c|c|c}
    \toprule[1.5pt]
    Comparison & Mean diff(pp) & 95\% CI (pp)  & $p_t$ & $p_{Holm}$ & $d_z$ & $p_{Wilcox}$ \\ \hline
    BGAT vs Fixed & 18.96 & [7.78, 30.15] & 0.0036 & 0.0134 & 1.14  & 0.005 \\ 
    BGAT vs Mean & 1.45  & [-0.03, 2.93] & 0.0533 & 0.0533 & 0.66  & 0.0674 \\ 
    BGAT vs K-Means & 2.75  & [0.85, 4.66] & 0.0093 & 0.0185 & 0.97  & 0.001 \\ 
    BGAT vs GMM-Int & 14.59 & [6.08, 23.10] & 0.0034 & 0.0134 & 1.15  & 0.0049 \\ 
    SUFF vs w/o SUFF & 2.8   & [1.33, 4.27] & 0.0017 & 0.0086 & 1.28  & 0.001 \\ 
    \bottomrule[1.5pt]
    \end{tabular}%
    }
    \caption{Paired two-sided \textit{t}-tests across 11 datasets ($n{=}11$). Entries are mean HOS differences in percentage points (pp); 95\% CIs are \textit{t}-based confidence intervals for the mean difference. $p_t$ are
unadjusted; $p_{\text{Holm}}$ are Holm--Bonferroni adjusted across the five planned comparisons; $p_{\text{Wilcoxon}}$ are unadjusted two-sided signed-rank \textit{p}-values as a robustness check. Cohen's $d_z=\bar{\Delta}/s_{\Delta}$.}
  \label{tab:ablation_significance}%
\end{table}%

 \textbf{Ablation Study Significance Tests. } 
 As our method is training-free and performs direct inference on the test set, its performance is minimally affected by random seeds, with standard deviations nearly zero. We therefore treat datasets as the paired unit for significance analysis of the core ablation results. Using CLIP as the backbone, we evaluate across 11 tasks, including the four domains of OfficeHome, the VisDA test domain, and six VTAB datasets. We compare our proposed BGAT against direct CLIP-based scoring and four alternative thresholding strategies to demonstrate the effectiveness of BGAT. Additionally, we compare results with and without SUFF (using BGAT as the thresholding method) to validate the contribution of the SUFF module.

 For each comparison, we compute per-dataset HOS differences and perform a \emph{paired} \(t\)-test, reporting the mean difference, the \(t\)-based 95\% confidence interval (CI), and Holm--Bonferroni adjusted \(p\)-values; nonparametric Wilcoxon results point in the same direction. Aggregate statistics are in Table~\ref{tab:ablation_significance}: BGAT vs.\ Fixed shows +18.96\,pp (95\% CI [7.78, 30.15], \(p_{\text{H}}=0.0134\)); BGAT vs.\ K-Means +2.75\,pp ([0.85, 4.66], \(p_{\text{H}}=0.0185\)); BGAT vs.\ GMM-Int +14.59\,pp ([6.08, 23.10], \(p_{\text{H}}=0.0134\)); BGAT vs.\ MEAN shows a +1.45\,pp positive, near-threshold trend at \(\alpha=0.05\); adding SUFF on top of BGAT  yields a significant +2.80\,pp ([1.33, 4.27], \(p_{\text{H}}=0.0086\)). The forest plot (Figure~\ref{fig:statistical_significance}) highlights comparisons whose CIs lie entirely to the right of 0; effect sizes (Cohen’s \(d_z\)) are generally large (Table~B.1), indicating not only statistical but also practically meaningful gains.

\begin{algorithm}[ht]
\caption{VLM-OpenXpert  Algorithm}
\label{alg:open-set}
\begin{algorithmic}[1]
\Require Unlabeled target dataset \( D_t = \{x_i^t\}_{i=1}^{N_t} \), known class label set \( L_t = \{l_1, \ldots, l_C\} \) where \( l_c \) denotes the \( c \)-th class, along with an image encoder E and a text encoder G.
\Ensure Final classification predictions for \( D_t \)
\State Extract features: \( X_{\text{all}} \gets E(D_t) \), \( X_{\text{text}} \gets G(L_t) \) \Comment{(Via Eq. \ref{equa:x_all_and_x_text})}
\State Obtain initial predictions: \( (\hat{y},\, \text{Scores}) \gets \text{Classify}(X_{\text{all}},\, X_{\text{text}}) \) \Comment{(Via Eqs. \ref{euqa:hat_y} and \ref{euqa:score_entropy})}
\State Normalize scores: \( \text{Scores} \gets \text{Box-CoxTransform}(\text{Scores}) \) \Comment{(Via Eqs. \ref{equa:estimate_lamda} and \ref{euqa:box_cox_trans})}
\State Estimate parameters: \( \mu_{\text{know}},\, \mu_{\text{unk}},\, T^*_{\text{init}} \gets \text{fit GMM}(\text{Scores}) \) \Comment{(Using Eq. \ref{euq:EM_GMM})}
\State Revert transformation: \( (\mu_{\text{know}},\, \mu_{\text{unk}},\, T^*_{\text{init}}) \gets \text{InverseBoxCox}(\cdot) \) \Comment{(Via Eq. \ref{euqa:inverse_box_cox})}
\State Partition features:
  \begin{itemize}
  \item \( X_{\text{know}} \gets \{ x \in X_{\text{all}} \mid \text{Score}(x) < \frac{1}{2}(\mu_{\text{know}} + T^*_{\text{init}}) \} \)
  \item \( X_{\text{unk}} \gets \{ x \in X_{\text{all}} \mid \text{Score}(x) > \frac{1}{2}(\mu_{\text{unk}} + T^*_{\text{init}}) \} \)
  \end{itemize}
\State Subspace decomposition:
  \begin{itemize}
  \item \( Z_{\text{know}} \gets \text{SVD}(X_{\text{know}}) \), \( Z_{\text{unk}} \gets \text{SVD}(X_{\text{unk}}) \) \Comment{(Via Eqs. \ref{euqa:svd_x_know}, \ref{svd_x_unk})}
  \item \( X_{\text{all}}^{\text{know}} \gets \text{Project}(X_{\text{all}},\, Z_{\text{know}}) \) \Comment{(Via Eq. \ref{euqa:proj_x_know})}
  \item \( X_{\text{all}}^{\text{unk}} \gets \text{Project}(X_{\text{all}},\, Z_{\text{unk}}) \) \Comment{(Via Eq. \ref{euqa:proj_x_unk})}
  \end{itemize}
\State Compute feature ratio: \( \alpha(X_{\text{all}}) \gets \frac{\|X_{\text{all}}^{\text{unk}}\|}{\|X_{\text{all}}\|} \) \Comment{(Via Eq. \ref{euqa:calculate_alpha})}
\State Filter features: \( X_{\text{all}}^* \gets \text{FeatureFilter}(X_{\text{all}},\, \alpha(X_{\text{all}})) \) \Comment{(Via Eq. \ref{euqa:sub_unkonw_feature})}
\State Recalculate scores and estimate the threshold \(T^*\):
  \begin{itemize}
  \item \( \text{Scores}^* \gets \text{RecomputeScores}(X_{\text{all}}^*,\, X_{\text{text}}) \)
  \item Execute steps 3, 4, and 5 again using \(\text{Scores}^*\) to obtain the estimated final threshold \(T^*\).
  \end{itemize}
\State Generate final predictions: \( \text{Predictions} \gets \text{Threshold-based Classify}(\text{Scores}^*,\, T^*,\, \hat{y}) \) \Comment{(Via Eq. \ref{euqa:pred_by_T})}
\Return \( \text{Predictions} \)
\end{algorithmic}
\end{algorithm}

\subsection{Case Study}
\label{appendix:Case-Study}
To analyze the tendency of removing certain unknown classes on the known classes, we visualize the distribution of the top 25 classes with the strongest tendency in the Office-Home dataset across four domains before and after filtering by the SUFF module. We also compare the change in entropy of all samples before and after filtering by the SUFF module.

As shown in Figure~\ref{fig:appendix_detail_confusion}, each cell represents the distribution of predictions for all samples of an unknown class in one row, with each column indicating the distribution of unknown classes predicted as a specific known class. The darker the cell color, the more samples from the unknown class are classified into the corresponding known class. As shown in the first row of Figure~\ref{fig:appendix_detail_confusion}, it is evident that in the original CLIP predictions, many unknown classes exhibit a strong tendency toward known classes, causing these samples to have scores very close to known classes. This results in difficulty distinguishing these samples during unknown class detection. As shown in the second row of Figure~\ref{fig:appendix_detail_confusion}, after filtering by the SUFF module, the number of unknown classes with significant tendencies is notably reduced, allowing these hard-to-detect unknown class samples to be separated from the known classes.

Additionally, as shown in the first row of Figure~\ref{fig:appendix_detail_entropy_his}, the entropy of unknown classes computed from the original CLIP features is spread across different score ranges and significantly overlaps with that of the known classes. However, as shown in the second row of Figure~\ref{fig:appendix_detail_entropy_his}, after filtering by the SUFF module, the entropy of the unknown classes shifts significantly toward the maximum value, greatly enhancing the separability between known and unknown classes. Furthermore, it can be observed that the entropy of the known classes does not shift to the right during the feature filtering process, indicating that the SUFF module does not destroy the separability of known class features while filtering unknown class features.

\section{Algorithm Details and Pseudocode}
In this section, we provide the detailed pseudocode of \ourmodel. Our method aims to detect unknown samples and classify known-class samples in an unlabeled target dataset without requiring any source-domain data or additional training. The approach consists of the following key steps: (1) We leverage a VLM (e.g., CLIP) as an encoder to perform zero-shot classification, obtaining the initial classification results for known classes and computing sample scores based on these results; (2) The proposed BGAT module estimates the distribution of sample scores and derives the estimated mean scores for known and unknown classes, along with an initial threshold; (3) Based on these estimates, we perform an initial selection of samples and apply the proposed SUFF module to filter out unknown-class features;(4) We then recompute sample scores using the filtered features and re-estimate the final threshold via the BGAT module; (5) Finally, we classify all samples using the refined scores, the final threshold, and the initial classification results. The full procedure is detailed in Algorithm \ref{alg:open-set}.

\end{document}